\documentclass[letterpaper, 10 pt, conference]{ieeeconf}  

\IEEEoverridecommandlockouts                              

\usepackage{tabularx, amsmath, amsfonts, graphicx, subcaption, booktabs, multirow, multicol, makecell, wrapfig, color, soul, dblfloatfix, balance, hyperref}

\newenvironment{small_ind_s_itemize}{\begin{list}{$\bullet$}
{\setlength{\rightmargin}{0em}
\setlength{\leftmargin}{2em}
\setlength{\itemsep}{0em}
\setlength{\itemindent}{-0.6em}
\setlength{\topsep}{0em}
\setlength{\parsep}{0em}}}{\end{list}}

\newcounter{ctr}

\newenvironment{small_ind_enumerate}{\begin{list}{\thectr.}
{\usecounter{ctr}
\setlength{\rightmargin}{\rightmargin}
\setlength{\leftmargin}{2em}
\setlength{\topsep}{\topsep}
\setlength{\itemindent}{-0.8em}
\setlength{\parsep}{\parsep}}}{\end{list}}

\title{\LARGE \bf
Continual Learning of\\ Knowledge Graph Embeddings
}

\author{Angel Daruna$^{1}$, Mehul Gupta$^{1}$, Mohan Sridharan$^{2}$ and Sonia Chernova$^{1}$
\thanks{$^{1}$Georgia Institute of Technology, Atlanta, GA. Email: {\tt\small \{adaruna3, mgupta320, chernova\}@gatech.edu}}
\thanks{$^{2}$University of Birmingham, Birmingham, UK. Email: {\tt\small m.sridharan@bham.ac.uk}}
\thanks{This work is supported in part by NSF IIS 1564080, NSF GRFP DGE-1650044, and ONR N00014-16-1-2835. Mohan Sridharan was supported in part by ONR N00014-17-1-2434. Any opinions, findings, and conclusions or recommendations expressed in this material are those of the author(s) and do not necessarily reflect the views of the supporters.}
}

\begin{document}

\maketitle
\thispagestyle{empty}
\pagestyle{empty}

\begin{abstract}
In recent years, there has been a resurgence in methods that use distributed (neural) representations to represent and reason about semantic knowledge for robotics applications. However, while robots often observe previously unknown concepts, these representations typically assume that all concepts are known a priori, and incorporating new information requires all concepts to be learned afresh. Our work relaxes this limiting assumption of existing representations and tackles the incremental knowledge graph embedding problem by leveraging the principles of a range of continual learning methods. Through an experimental evaluation with several knowledge graphs and embedding representations, we provide insights about trade-offs for practitioners to match a semantics-driven robotics applications to a suitable continual knowledge graph embedding method.

\end{abstract}

\section{Introduction}
\label{sec:intro}
Representing and reasoning about semantic knowledge is a key task in robotics. In recent years, there has been a resurgence in methods that use distributed (neural) representations, e.g., word and knowledge graph embeddings, for this task in the context of navigation~\cite{fulda2017harvesting}, grounding~\cite{thomason2018guiding}, affordance modeling~\cite{daruna2019robocse}, success detection~\cite{scalise2019improving}, manipulation~\cite{paulius2020motion}, and instruction following~\cite{arkin2020multimodal}. While robots frequently observe previously unknown concepts, these embedding algorithms typically assume that all embedding concepts are known a priori, and incorporating new information requires all concepts to be learned afresh. In addition, in robotics applications, the limited availability of computational resources and storage, and concerns regarding storing sensitive information, can make batch learning with all observed data infeasible. We seek to relax this static assumption in knowledge graph embedding and enable adaptive revision of distributed representation of semantic knowledge for robots. 

Towards achieving our objective, we draw on \textit{Continual Learning}, the research area which focuses on the challenging problem of incrementally revising learned neural representations~\cite{parisi2019continual}. Existing continual learning methods have predominantly been applied to object recognition and include regularization~\cite{zenke2017continual,kirkpatrick2017overcoming}, architecture modification~\cite{rusu2016progressive,lomonaco2017core50}, generative replay~\cite{shin2017continual,van2018feedback}, and a reformulation of regularization for knowledge graph embedding~\cite{song2018enriching}. However, continual learning methods remain largely unexplored for knowledge graph embedding. Furthermore, the implications of any related assumptions for robotics is not well documented because existing methods focus on the final inference performance and define different task specific measures~\cite{diaz2018don}.

Our work makes three contributions. First, we reformulate and extend the underlying principles of five representative continual learning methods: (i) Progressive Neural Networks~\cite{rusu2016progressive}; (ii) Copy Weight Re-Init~\cite{lomonaco2017core50}; (iii) L2 regularization~\cite{kirkpatrick2017overcoming}; (iv) Synaptic Intelligence~\cite{zenke2017synaptic}; and (v) Deep Generative Replay~\cite{shin2017continual}, and apply them to the \textit{continual knowledge graph embedding} (CKGE) problem. Second, we introduce an empirically evaluated heuristic sampling strategy to generate CKGE datasets from knowledge graphs, since benchmark datasets do not exist for the CKGE problem. Third, we build on existing continual learning measures~\cite{lesort2020cl_robotics} to characterize the use of each reformulated method for robot tasks that leverage semantic knowledge.

For evaluation, we consider two knowledge graph embedding representations with different assumptions and loss functions: TransE~\cite{bordes2013translating} and Analogy~\cite{liu2017analogical}; and three benchmark knowledge graphs (WN18RR, FB15K237~\cite{dettmers2018convolutional}, and AI2Thor~\cite{daruna2019robocse}). We also evaluated each adapted method under unconstrained, data-constrained, and time-constrained settings by sampling from a knowledge graph used in prior robotics work~\cite{daruna2019robocse}, containing actions, locations, objects, and other concepts. Experimental results indicate that: (i) our generative replay approach outperforms other methods; (ii) there are interesting trade-offs between inference capability, learning speed, and memory usage that should be considered when choosing a CKGE method; and (iii) insights gained from exploring these trade-offs enable us to select a CKGE method that best matches the constraints of a given robotics application that models semantic knowledge.

\section{Related Work \& Background}
\label{sec:related}
We motivate our contributions by discussing background information and related work.

\textbf{Modeling Semantic Knowledge in Robotics} is often achieved using an explicit model of world semantics in the form of a knowledge graph $\mathcal{G}$ composed of individual facts or triples $(h,r,t)$; $h$ and $t$ are the head and tail entities (respectively) for which the relation $r$ holds, e.g., {$($\textit{cup, hasAction, fill}$)$}~\cite{beetz2018know,chernovasituated,saxena2014robobrain,zhu2014reasoning}.  Recent work has modeled $\mathcal{G}$ using distributed representations because of their ability to approximate proximity of meaning from vector computations~\cite{fulda2017harvesting, thomason2018guiding,daruna2019robocse,scalise2019improving,paulius2020motion,arkin2020multimodal}.

\textbf{Multi-relational (knowledge graph) embeddings} are distributed representations that model $\mathcal{G}$ in vector space~\cite{wang2017kge_survey}, learning a continuous vector representation from a dataset of triples $\mathcal{D}\!=\!\big\{(h,r,t)_i,y_i|\,h_i,t_i\!\in\!\mathcal{E},r_i\!\in\!\mathcal{R},y_i\!\in\!\{0,1\}\big\}$, with $i\!\in\!\{1...|\mathcal{D}|\}$. Here $y_i$ denotes whether relation $r_i \in \mathcal{R}$ holds between $h_i, t_i \in \mathcal{E}$. Each entity $e\!\in\!\mathcal{E}$ is encoded as a vector $\textbf{v}_e\!\in\!\mathbb{R}^{d_\mathcal{E}}$, and each relation $r\!\in\!\mathcal{R}$ is encoded as a mapping between vectors $\textbf{W}_r\!\in\!\mathbb{R}^{d_\mathcal{R}}$, where $d_\mathcal{E}$ and $d_\mathcal{R}$ are the dimensions of vectors and mappings respectively~\cite{wang2017kge_survey,nickel2016review}. The embeddings for $\mathcal{E}$ and $\mathcal{R}$ are typically learned using a scoring function $f(h,r,t)$ that assigns higher (lower) values to positive (negative) triples~\cite{nickel2016review}. The learning objective is thus to find a set of embeddings $\Theta = \big\{\{\textbf{v}_{e}|\,e\in\mathcal{E}\},\{\textbf{W}_{r}|\,r\in\mathcal{R}\}\big\}$ that minimizes the loss $\mathcal{L}_\mathcal{D}$ over $\mathcal{D}$. Loss $\mathcal{L}_\mathcal{D}$ can take many forms depending on the multi-relational embedding representation used, e.g., Margin-Ranking Loss~\cite{bordes2013translating} or Negative Log-Likelihood Loss~\cite{liu2017analogical}. However, all entities and relations are assumed to be known before training~\cite{wang2017kge_survey,nickel2016review}, which may be infeasible for robots observing new concepts or new facts about existing concepts.
 
\textbf{Continual learning} has evolved as a subarea of life-long machine learning. It focuses on neural networks and seeks to learn new domains, classes, or tasks over time without forgetting previously learned knowledge~\cite{parisi2019continual}. Methods proposed in the narrow context of object recognition include regularization~\cite{zenke2017continual,kirkpatrick2017overcoming}, architecture modification~\cite{rusu2016progressive,lomonaco2017core50}, and generative replay~\cite{shin2017continual,van2018feedback}. We explore and adapt five representative methods~\cite{kirkpatrick2017overcoming,rusu2016progressive,lomonaco2017core50,shin2017continual,zenke2017synaptic}. Different categories of continual learning scenarios exist in the literature based on whether there are shifts in the input or output distributions, and whether the inputs and outputs share the same representation space~\cite{hsu2018re}. Among existing categories, we chose \textit{Incremental Class Learning}~(ICL) because it best matches the assumptions of robot systems representing semantic knowledge, with the distribution of input data and target labels changing across learning sessions as the robot incrementally observes disjoint sets of new facts about new and existing concepts.

In CKGE, the dataset $\mathcal{D}$ of a knowledge graph $\mathcal{G}$ is split into multiple datasets $\mathcal{D}^{n}$ where $n$ indicates the learning session~\cite{song2018enriching}. Each $\mathcal{D}^{n}$ contains a disjoint set of all triples of a subset of entities and relations. For a robot observing new facts, the size of the set of observed entities, relations, and triples grows, (e.g., $|\mathcal{E}^{n}| \leq |\mathcal{E}^{n+1}|$), and the embedding must consider new facts and concepts in each learning session. In such a learning scenario, the objective is to find a set of embeddings $\Theta^n = \big\{\{\textbf{v}^n_{e}|\,e\in\mathcal{E}^n\},\{\textbf{W}^n_{r}|\,r\in\mathcal{R}^n\}\big\}$ that minimize the loss $\mathcal{L}_{\mathcal{D}^n}$ over the dataset for all time steps. Of the range of continual learning methods, only L2-regularization has been applied to CKGE~\cite{song2018enriching}; more sophisticated methods that have shown promise in other domains, e.g., generative replay, remain unexplored. Also, important measures for robotics, such as learning efficiency and model complexity, are not well documented for representative techniques~\cite{lesort2020cl_robotics}, making it difficult to evaluate the suitability of these methods for modeling semantic knowledge in robotics. Our work is designed to fill these gaps.

\textbf{Dynamic Graph Embedding} is a related approach focused on modeling dynamic graphs with applications to social networks, biology, computational finance, and other domains~\cite{kazemi2020representation}. A work from~\cite{kazemi2020representation} on dynamic \textit{knowledge} graph embedding~\cite{wu2019efficiently} assumes the knowledge graph in each time step is complete, and models the changes in the knowledge graph across learning sessions taking as inputs the knowledge graph from the prior and current learning session. Our work uses a different set of assumptions because we model a scenario where a robot is observing disjoint subsets of a complete knowledge graph. We view our approach as a variant of dynamic knowledge graph embedding in which only a subset of the complete knowledge graph is available for training each learning session. As a result, our algorithms need to take into account problems of catastrophic forgetting. Catastrophic forgetting~\cite{parisi2019continual} occurs when a neural representation that was optimized for a prior dataset is trained with new dataset. The neural network's weights are tuned to the new dataset, resulting in a potential loss in performance for classes and tasks not included in the new dataset.

\section{Continual Knowledge Graph Embedding}
\label{sec:approach}
We seek to characterize the use of continual learning methods for knowledge graph embedding in robotics by exploring the associated assumptions and trade-offs. In this section, we describe how we reformulate and extend the principles of five carefully selected representative continual learning techniques to develop \textit{continual knowledge graph embedding} (CKGE) methods. These methods were designed for traditional neural networks and required varying levels of innovation to support knowledge graph embeddings. In each case, we carefully considered the suitability of its principles to support the desired capabilities and assumptions of knowledge graph embeddings.

\subsection{Architectural Modification Methods}
\label{sec:approach-archmod}
Among the methods that modify the architecture of a neural network to accommodate new training data while minimizing performance losses over older data, we adapted two methods for knowledge graph embeddings.

\textbf{Progressive Neural Networks (PNN)}~\cite{rusu2016progressive} add copies of existing layers of a multi-layered neural network for each new learning session. When a new learning session begins, existing weights are frozen so that back-propagated gradients do not affect the performance over data from previous sessions. Also, lateral connections are made between successive layer copies to enable the forward transfer of previously learned weights. To make PNNs applicable to knowledge graph embedding, we first expand the embedding matrices $\textbf{v}^{n}\!\in\!\mathbb{R}^{|\mathcal{E}^{n}| \times d_\mathcal{E}}$ and $\textbf{W}^{n}\!\in\!\mathbb{R}^{|\mathcal{R}^{n}| \times d_\mathcal{R}}$ to include new entities and relations in the learning session $n$. Second, we freeze embeddings for entities and relations encountered in prior learning sessions to prevent their corruption in the current learning session. Instead of creating separate copies of these embedding matrices for each learning session, we only expand the existing matrices to promote forward transfer of prior embeddings in new learning sessions.

\textbf{Copy Weight with ReInit (CWR)}~\cite{lomonaco2017core50} maintains the weights of the final layer of the network during a new learning session (i.e. temporary weights, TW), separate from the corresponding weights trained in prior learning sessions (i.e. consolidated weights, CW) to avoid corruption. Other than the two sets of final layer weights considered during (continual) learning, the weights of other layers are frozen and shared across learning sessions. TW are re-sized and re-initialized in each learning session according to the number of classes being trained. After each learning session, the TW for new classes are copied over to CW, which acts as a memory buffer separate from the network. If a previously trained class is encountered, relevant entries in TW are averaged with those in CW. Training for the subsequent session begins by re-sizing and re-initializing TW.

To apply the principles of CWR to knowledge graph embedding, we first introduce two sets of embeddings: consolidated embeddings (CE) \{$\textbf{v}^{n}_{ce}$,$\textbf{W}^{n}_{ce}$\} and temporary embeddings (TE) \{$\textbf{v}^{n}_{te}$,$\textbf{W}^{n}_{te}$\}. Second, for each learning session, we resize and re-initialize the TE for entities $\textbf{v}^{n}_{te}$ and relations $\textbf{W}^{n}_{te}$ based on the number of entities and relations (respectively) in the session. After the session, we move TE into CE by copying new embeddings or averaging existing ones. As a result, the number of CE increases monotonically in each learning session with the number of observed entities $\mathcal{E}^{n}$ and relations $\mathcal{R}^{n}$ so that $\textbf{v}^{n}_{ce}\!\in\!\mathbb{R}^{|\mathcal{E}^{n}| \times d_\mathcal{E}}$ and $\textbf{W}^{n}_{ce}\!\in\!\mathbb{R}^{|\mathcal{R}^{n}| \times d_\mathcal{R}}$) (respectively); the number of TE changes in each learning session according to the number of entities and relations in that learning session's dataset $\mathcal{D}^{n}$.

\subsection{Regularization Methods}
\label{sec:approach-regularization}
Freezing previously learned weights prevents their corruption in subsequent sessions, but also prevents shared weights from being revised to better accommodate new concepts. Some continual learning methods allow adjustments to shared weights that perform well for prior and new sessions; they do so by enforcing some regularization terms in new learning sessions. We reformulate two such approaches for knowledge graph embeddings.

\textbf{L2 Regularization (L2R)}~\cite{kirkpatrick2017overcoming,song2018enriching,hsu2018re} is adapted in our approach by adding a regularization term to the learning session loss $\mathcal{L}_{\mathcal{D}^{n}}$, encouraging the trained weights to not deviate from their previous values: 
\begin{equation}
    \small \mathcal{L}_{\mathcal{D}^{n}} + \lambda \cdot
    \Big{(}||\textbf{v}^{n}_{e}-\textbf{v}^{n-1}||^{2}_{2} + ||\textbf{W}^{n}_{r}-\textbf{W}^{n-1}||^{2}_{2}\Big{)}
    \label{eq:l2r}
\end{equation}
where $e\!\in\!\mathcal{E}^{n-1}$, $r\!\in\!\mathcal{R}^{n-1}$, and $\lambda$ is a regularization scaling term tuned as a hyper-parameter. L2R can be rather strict because it penalizes all dimensions of an embedding equally, whereas a subset of the embedding dimensions often contribute more to loss or predictive abilities than others.

\textbf{Synaptic Intelligence (SI)}~\cite{zenke2017synaptic} extends L2R by considering the weight-specific contributions to the reduction in loss over a learning session. These contributions are quantified by summing the gradients that each weight adjustment contributes to the loss and using the total loss reduction as a normalizer. SI is generic enough to apply to knowledge graph embeddings with minimal changes because it is formulated in terms of the weight and loss trajectories. Equation~\ref{eq:si} defines our implementation of SI for knowledge graph embedding, re-using terms from~\cite{zenke2017synaptic}:
\begin{equation}
    \small \mathcal{L}_{\mathcal{D}^{n}} + \lambda \cdot
    \Big{(}||\Omega_{e}(\textbf{v}^{n}_{e}-\textbf{v}^{n-1})||^{2}_{2} + ||\Omega_{r}(\textbf{W}^{n}_{r}-\textbf{W}^{n-1})||^{2}_{2}\Big{)}
    \label{eq:si}
\end{equation}
where $e\!\in\!\mathcal{E}^{n-1}$, $r\!\in\!\mathcal{R}^{n-1}$, $\Omega$ is the parameter regularization strength~\cite{zenke2017synaptic}, and $\lambda$ is a regularization scaling term tuned as a hyper-parameter for a particular representation. Elastic Weight Consolidation (EWC)~\cite{kirkpatrick2017overcoming} was also considered but not used because the assumptions made by the Fisher Information matrix of EWC are not satisfied by many knowledge graph embeddings, e.g., those using Margin-Ranking Loss.

\vspace{-0.2em}
\subsection{Generative Replay Methods}
\vspace{-0.5em}
\label{sec:approach-replay}

\begin{wrapfigure}{r}{0.2\textwidth}
    \vspace{-1cm}
	\centering 
	\includegraphics[width=0.2\textwidth]{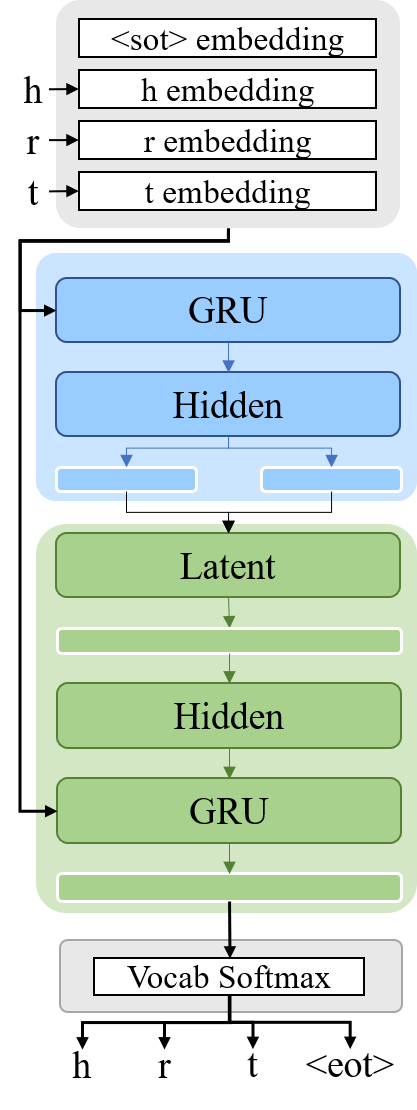}
	\vspace{-0.7cm}
	\caption[]{\small DGR architecture. Layers with white outline are linear.}
	\label{fig:dgr_arch}
	\vspace{-0.5cm}
\end{wrapfigure}

Instead of maintaining model weights across learning sessions, generative replay methods learn generative models of the distribution of training data from previous learning sessions. Then, batch learning is approximated by sampling from the learned distribution and the training data from the current learning session. We reformulate one such method for knowledge graph embeddings.

\textbf{Deep Generative Replay (DGR)} \cite{shin2017continual,van2018feedback} is a continual learning method that uses a generative model $\textbf{G}$ to approximate the distribution of all observed training examples (i.e. $\mathcal{D}$), and trains a discriminative model (i.e., solver) to perform a task. In the initial learning session, generator $\textbf{G}^{0}$ and solver are trained using examples in $\mathcal{D}^{0}$. In any subsequent learning session $i$, a new generator $\textbf{G}^{i}$ and solver are trained using examples in $\mathcal{D}^{i}$ and samples from $\textbf{G}^{i-1}$ that approximate $\mathcal{D}^{i-1}$, thus approximating training with $\mathcal{D}^{i-1} \cup \mathcal{D}^{i}$.

The challenge in applying the principles of DGR to knowledge graph embeddings is designing an effective generator, as the solver is determined by the representation used, i.e., $\Theta = \big\{\{\textbf{v}_{e}|\,e\in\mathcal{E}\},\{\textbf{W}_{r}|\,r\in\mathcal{R}\}\big\}$). Sampling training examples is a known problem in knowledge graph embedding\footnote{Others have used GANs to generate negative examples~\cite{wang2018incorporating,cai2017kbgan,dai2020generative}, but we cannot use these methods because their generators require positive examples as input.}, but prior work has shown that a Variational Auto-Encoder (VAE) can be used to sample sequences of discrete tokens~\cite{bowman2015seqvae}. We treat each triple as a sequence of discrete tokens to design our VAE-based generator. 

Figure~\ref{fig:dgr_arch} shows our VAE architecture that uses Gated-Recurrent Units to encode and decode the triples to and from the latent space $z$. Input triples $(h,r,t)$ to the encoder are first transformed into token embedding sequences $x=(\nu_{h},\nu_{r},\nu_{t})$, where $\nu\!\in\!\mathbb{R}^{|\mathcal{E}^{n}| + |\mathcal{R}^{n}| \times d_\mathcal{V}}$ is a token embedding learned by the encoder with dimensionality $d_\mathcal{V}$. The encoder, shown in blue in Figure~\ref{fig:dgr_arch}, is a learned posterior recognition model $q(z|x)$ that approximates the posterior distribution over $z$, conditioned on the input triple sequences $x$. Unlike a standard auto-encoder, the encoder is encouraged to keep the learned posterior $q(z|x)$ close to the prior over the latent space $p(z)$, which is a standard Gaussian. A similarity constraint based on the KL divergence measure in the objective function allows samples to be generated from the latent space. These samples are decoded using the decoder, shown in green in Figure~\ref{fig:dgr_arch}, to maximize $p(x|z)$, the likelihood of a triple sequence $x$ conditioned on its encoded latent space vector $z$, as in a standard auto-encoder. The output sequences of the VAE are transformed back into a triples using a Softmax function over all tokens (i.e., $e\!\in\!\mathcal{E}^{n}$ and $r\!\in\!\mathcal{R}^{n}$). The objective function for this architecture is:
\vspace{-0.3em}
\begin{equation}
    \small -\textrm{KL}\big(q(z|x) || p(z)\big) \cdot \alpha(\textrm{epoch}) + \mathbb{E}_{q(z|x)}\big[\textrm{log}\,p(x|z)\big]
    \vspace{-0.3em}
    \label{eq:dgr}
\end{equation}
where an additional term $\alpha(\cdot)$ is included to anneal the KL divergence loss, preventing issues such as vanishing gradients caused by posterior sampling and KL divergence loss terms being driven to zero~\cite{bowman2015seqvae}. $\alpha(\cdot)$ is a function of the number of epochs trained for the learning session:
\vspace{-0.3em}
\begin{equation}
    \small \alpha(\textrm{epoch}) = \frac{\lambda_{am}}{1 + e^{-\lambda_{as} \big(\textrm{epoch}\,-\,\lambda_{ap}\big)}}
    \vspace{-0.3em}
    \label{eq:anneal}
\end{equation}
where $\lambda_{am}$, $\lambda_{as}$, and $\lambda_{ap}$ are hyper-parameters tuned during training to control the maximum value, slope, and position of the annealing function, respectively.

\section{Experimental Setup}
\label{sec:exp_pro}
We evaluate our CKGE methods on two multi-relational embedding representations: TransE~\cite{bordes2013translating} and Analogy~\cite{liu2017analogical}; and three benchmark knowledge graphs: AI2Thor~\cite{daruna2019robocse}, FB15K237~\cite{dettmers2018convolutional}, and WN18RR~\cite{dettmers2018convolutional}. The last two knowledge graphs are challenging and have been widely used in the graph embedding literature~\cite{dettmers2018convolutional,cai2017kbgan,das2017go}. AI2Thor contains relations and entities related to service robotics, e.g., locations of objects, actions that can be performed on objects, and the outcomes that result from these actions~\cite{daruna2019robocse}. We report the accuracy and complexity of each method based on seven performance measures chosen from prior continual learning work in robotics~\cite{lesort2020cl_robotics}. In each trial, the evaluation task is triplet prediction, a fundamental knowledge graph embedding task~\cite{fulda2017harvesting,daruna2019robocse} with a well-defined experimental setup~\cite{bordes2013translating,wang2017kge_survey} as described later in this section.

\textbf{CKGE datasets:} Since there is no established benchmark dataset for CKGE, we introduce three standard evaluation datasets that we obtain by sampling. Our heuristic sampling strategy emulates the New Instances and Concepts scenario presented in~\cite{lesort2020cl_robotics} under the categorization of the nature of data samples within training sets. Therefore, our sampling strategy models the scenario where a robot explores a world and discovers new triple instances that contain new concepts (i.e. entities or relations), new triple instances that contain previously observed concepts, and triple instances that have been previously observed. Consider a knowledge graph $\mathcal{G}$ whose triples $\mathcal{D}$ have been split into a training set $\mathcal{D}_{Tr}$, validation set $\mathcal{D}_{Va}$, and test set $\mathcal{D}_{Te}$. Our approach for generating  datasets for $n=\{1,...,N\}$ learning sessions is:
\begin{small_ind_enumerate}
    \item \textit{Sample training triples}: uniformly sample without replacement $\frac{|\mathcal{D}_{Tr}|}{N}$ triples from training set $\mathcal{D}_{Tr}$ of $\mathcal{G}$. These triples form training dataset $\mathcal{D}^{n}_{Tr}$.
   
    \item \textit{Extract entities and relations}: create a set of entities $E^{n}$ and a set of relations $R^{n}$ for this session from the triples in $\mathcal{D}^{n}_{Tr}$. The set of all observed entities (relations), i.e., $\mathcal{E}^{n}$ ($\mathcal{R}^{n}$) is the union of current and prior $E^{n}$ ($R^{n}$).
   
    \item \textit{Construct $n^{th}$ validation and test sets}: extract from $\mathcal{D}_{Va}$ and $\mathcal{D}_{Te}$ the triples whose head, relation, and tail belong to $E^{n}$ and $R^{n}$ (respectively). These triples form validation set $\mathcal{D}^{n}_{Va}$ and test set $\mathcal{D}^{n}_{Te}$ of the $n^{th}$ session.
   
    \item \textit{Remove sampled training triples}: remove $\mathcal{D}^{n}_{Tr}$ from $\mathcal{D}_{Tr}$ of $\mathcal{G}$. 
    
    \item Repeat steps 1-4 until no training triples exist in $\mathcal{G}$ or a predefined number of iterations are completed.
\end{small_ind_enumerate}

\begin{table}
    \setlength{\tabcolsep}{4pt} 
    \vspace{0.1cm}
	\caption{\centering \small CKGE Datasets; Benchmarks}
	\vspace{-0.2cm}
    \resizebox{\columnwidth}{!}{
        \begin{tabular}{l|ccccc} \toprule
        & \multicolumn{5}{c}{WN18RR-5-LS} \\ \midrule
        & LS-1 & LS-2 & LS-3 & LS-4 & LS-5 \\ \midrule
        $|E^{n}|$ & 20,368/(50\%) & 20,389/(73\%) & 20,249/(87\%) & 20,463/(95\%) & 20,437/(99\%) \\ 
        $|R^{n}|$ & 11/(100\%) & 11/(100\%) & 11/(100\%) & 11/(100\%) & 11/(100\%) \\ 
        $|\mathcal{D}^{n}_{Tr}|$ & 17,367/(20\%) & 17,367/(40\%) & 17,367/(60\%) & 17,367/(80\%) & 17,367/(100\%) \\ 
        $|\mathcal{D}^{n}_{Va}|$ & 1,117/(37\%) & 1,141/(57\%) & 1,187/(71\%) & 1,190/(80\%) & 1,184/(86\%) \\ 
        $|\mathcal{D}^{n}_{Te}|$ & 1,168/(37\%) & 1,159/(57\%) & 1,218/(72\%) & 1,173/(81\%) & 1,175/(87\%) \\ \midrule
        & \multicolumn{5}{|c}{FB15K237-5-LS} \\ \midrule
        & LS-1 & LS-2 & LS-3 & LS-4 & LS-5 \\ \midrule
        $|E^{n}|$ & 13,143/(90\%) & 13,106/(96\%) & 13,115/(98\%) & 13,089/(99\%) & 13,163/(100\%) \\
        $|R^{n}|$ & 237/(100\%) & 237/(100\%) & 237/(100\%) & 237/(100\%) & 237/(100\%) \\ 
        $|\mathcal{D}^{n}_{Tr}|$ & 54,423/(20\%) & 54,423/(40\%) & 54,423/(60\%) & 54,423/(80\%) & 54,423/(100\%) \\ 
        $|\mathcal{D}^{n}_{Va}|$ & 17,013/(97\%) & 16,929/(99\%) & 16,917/(100\%) & 16,882/(100\%) & 16,905/(100\%) \\ 
        $|\mathcal{D}^{n}_{Te}|$ & 19,776/(97\%) & 19,727/(99\%) & 19,734/(99\%) & 19,758/(100\%) & 19,801/(100\%) \\ \bottomrule
        \end{tabular}
        \label{tbl:benchmark_datasets}
    }
    \vspace{0.1em}
    \caption{\centering \small CKGE Datasets; Robotics}
    \vspace{-0.2em}
    \resizebox{\columnwidth}{!}{
      \begin{tabular}{l|ccccc} \toprule
        & LS-1 & LS-2 & LS-3 & LS-4 & LS-5 \\ \midrule
        $|E^{n}|$ & 176/(84\%) & 175/(95\%) & 177/(97\%) & 171/(99\%) & 169/(99\%) \\
        $|R^{n}|$ & 11/(100\%) & 11/(100\%) & 11/(100\%) & 11/(100\%) & 11/(100\%) \\
        $|\mathcal{D}^{n}_{Tr}|$ & 17,367/(20\%) & 17,367/(40\%) & 17,367/(60\%) & 17,367/(80\%) & 17,367/(100\%) \\ 
        $|\mathcal{D}^{n}_{Va}|$ & 1,117/(37\%) & 1,141/(57\%) & 1,187/(71\%) & 1,190/(80\%) & 1,184/(86\%) \\ 
        $|\mathcal{D}^{n}_{Te}|$ & 1,168/(37\%) & 1,159/(57\%) & 1,218/(72\%) & 1,173/(81\%) & 1,175/(87\%) \\ \bottomrule
      \end{tabular}
      \label{tbl:robot_dataset}
    }
    \vspace{-0.6cm}
\end{table}

\noindent
We generated three CKGE datasets with $n=5$ sessions using our approach on two established benchmark knowledge graphs in the graph embedding community (WN18RR and FB15K237~\cite{dettmers2018convolutional}) and a knowledge graph used in robotics (AI2Thor~\cite{daruna2019robocse}). Tables~\ref{tbl:benchmark_datasets} and \ref{tbl:robot_dataset} report statistics of each dataset. The columns of the tables denote the learning session (LS-X, X$\in[1,5]$), while rows correspond to the statistics, e.g., $|E^{n}|$ is the size of the entity set. Individual cells indicate the value, with coverage with respect to the original knowledge graph shown in parentheses. For instance, in LS-2 of WN18RR-5-LS, there are $20,389$ entities and $73\%$ of all entities in WN18RR have been observed. Note that our sampling strategy empirically produces datasets with better coverage and higher percentages of new training triples each learning session, i.e., more challenging datasets for CKGE, than previous methods such as entity sampling~\cite{song2018enriching}. Furthermore, our sampling strategy makes the distribution of the $n$ training sets more closely match the original $\mathcal{D}_{Tr}$ than entity sampling by ensuring sampling without replacement \big($\mathcal{D}^{n}_{Tr}\,\bigcap\,\mathcal{D}^{n+1}_{Tr}\!=\!\emptyset\,\forall\,n$\big). In addition to our chosen sampling strategy, we repeated all experiments with two other sampling strategies, entity and relation sampling, detailed in the Appendix.

\textbf{Evaluation procedure:} The evaluation task is to predict complete triplets from incomplete ones in test splits $\mathcal{D}^{n}_{Te}$, i.e., predict $h$ given $(r, t)$ or $t$ given $(h, r)$. To perform triplet prediction, each test triplet $(h,r,t)$ is first corrupted by replacing the head (or tail) entity with every other possible entity in the current session $\mathcal{E}^{n}$. Then, to avoid underestimating the embedding performance, we remove all corrupted test triplets that still represent a valid relationship between the corresponding entities; this is known as the “filtered” setting in the literature~\cite{bordes2013translating}. Last, scores are computed for each test triplet and its (remaining) corrupted triplets using the scoring function $f(h, r, t)$ (defined below), then ranked in descending order. 

Recall that we consider two knowledge graph embedding representations to show the generality of our methods: TransE and Analogy. TransE represents relationships as translations between entities, i.e., $\textbf{v}_{h}+\textbf{W}_{r}=\textbf{v}_{t}$~\cite{bordes2013translating}. It uses the scoring and margin ranking loss functions in Equations~\ref{eq:transe_score} and~\ref{eq:transe_loss}, where $[x]_{+} = max(0, x)$, $\gamma$ is the margin, and $(h',r,t')$ are corrupted triples in a corrupted knowledge graph $\mathcal{G}'$. Embeddings are subject to normalization constraints (i.e. $||\textbf{v}_{e}||_{2}\leq1\,\forall\,e\in\mathcal{E}$ and $||\textbf{W}_{r}||_{2}\leq1\,\forall\,r\in\mathcal{R}$) to prevent trivial minimization of $\mathcal{L}$ by increasing entity embedding norms during training.
\vspace{-0.5em}
\begin{equation}
    \small f(h,r,t) = ||\textbf{v}_{h}+\textbf{W}_{r}-\textbf{v}_{t}||_{1}
    \vspace{-0.5em}
    \label{eq:transe_score}
\end{equation}
\vspace{-0.5em}
\begin{equation}
    \small \mathcal{L} = \sum\limits_{\substack{\small (h,r,t)\in\mathcal{G}, \\ \small (h',r,t')\in\mathcal{G}'}}
    [f(h,r,t) + \gamma - f(h',r,t')]_{+}
    \vspace{-0.5em}
    \label{eq:transe_loss}
\end{equation}
Analogy, on the other hand, represents relationships as (bi)linear mappings between entities, i.e., $\textbf{v}^{\top}_{h}\textbf{W}_{r}=\textbf{v}^{\top}_{t}$~\cite{liu2017analogical}. It uses the scoring and negative log loss functions in Equations~\ref{eq:analogy_score} and~\ref{eq:analogy_loss} where $\sigma$ is a sigmoid function, $y$ is a label indicating whether the triple is corrupted, and $\mathcal{G}'$ is the corrupted knowledge graph. Additionally, the linear mappings (i.e. relations) are constrained to form a commuting family of normal mappings, i.e., $\textbf{W}_{r}\textbf{W}^{\top}_{r}=\textbf{W}^{\top}_{r}\textbf{W}_{r}\,\forall\,r\in\mathcal{R}$ and $\textbf{W}_{r}\textbf{W}_{r'}=\textbf{W}_{r'}\textbf{W}_{r}\,\forall\,r,r'\in\mathcal{R}$, to promote analogical structure within the embedding space.
\vspace{-0.5em}
\begin{equation}
    \small f(h,r,t) = \langle\textbf{v}^{\top}_{h}\textbf{W}_{r},\textbf{v}_{t}\rangle
    \vspace{-0.5em}
    \label{eq:analogy_score}
\end{equation}
\vspace{-0.5em}
\begin{equation}
    \small \mathcal{L} = \sum\limits_{\small (h,r,t,y)\in\,{\mathcal{G},\mathcal{G'}}}
    -\textrm{log}\,\sigma(y \cdot f(h,r,t))
    \vspace{-0.5em}
    \label{eq:analogy_loss}
\end{equation}

\textbf{Evaluation measures:} We build on existing measures to characterize each of our CKGE methods. We consider different factors important for robotics applications modeling semantic knowledge, e.g., inference, memory usage, and learning efficiency. In addition to the only measure provided in prior CKGE work~\cite{song2018enriching} (i.e. inference performance), we report seven other robotics-oriented metrics cataloged in~\cite{lesort2020cl_robotics} that measure unique aspects of continual learning algorithms. Specifically, for inference performance, we consider the \textit{mean reciprocal rank} of correct triplets (MRR) and the proportion of the correct triplets ranked in the top 10 (Hits@10). During each learning session, we compute the evaluation measures for the test sets of all learning sessions to characterize the effect of learning on prior, current, and future learning sessions. During the $n^{th}$ learning session of $N$ total sessions, the two training-test inference performance matrices $\textbf{M} \in \mathbb{R}^{N \times N}$ (for MRR and Hits@10) are used to compute four measures that summarize accuracy and forgetting across learning sessions: (i) Average accuracy (ACC) measures the average accuracy across learning sessions---Equation~\ref{eq:acc}; (ii) Forward Transfer (FWT) measures zero-shot learning in future sessions by transferring weights learned in prior session(s)---Equation~\ref{eq:fwt}; (iii) Backwards Transfer (+BWT) measures the improvement over expected performance of a prior learning session as a result of learning in future sessions---Equation~\ref{eq:+bwt}; and (iv) Remembering (REM) measures how performance in a learning session degrades as a result of learning in subsequent sessions---Equation~\ref{eq:rem}.

\begin{multicols}{2}
    \vspace*{-1cm}
    \begin{equation}
        \small \textrm{\small ACC} = \frac{\sum_{i \geq j}^{N}\textbf{M}_{i,j}}{\frac{N(N+1)}{2}}
        \label{eq:acc}
    \end{equation}\break
    \vspace*{-1cm}
    \begin{equation}
        \small \textrm{\small FWT} = \frac{\sum_{i < j}^{N}\textbf{M}_{i,j}}{\frac{N(N-1)}{2}}
        \label{eq:fwt}
    \end{equation}
\end{multicols}
\vspace{-0.5cm}
\begin{equation}
    \small \textrm{\small BWT} = \frac{\sum_{i=2}^{N}\sum_{j=1}^{i-1}(\textbf{M}_{i,j}-\textbf{M}_{j,j})}{\frac{N(N-1)}{2}}
    \vspace{-0.7cm}
    \label{eq:bwt}
\end{equation}

\begin{multicols}{2}
    \begin{equation}
        \small \textrm{\small+BWT}\!=\!\textrm{max}(0, \textrm{BWT})
        \label{eq:+bwt}
    \end{equation} \break
    \begin{equation}
        \small \textrm{\small REM}\!=\!1\!-\!|\textrm{min}(0, \textrm{BWT})|\!
        \label{eq:rem}
    \end{equation}
\end{multicols}
\vspace{-0.3cm}

\noindent
Other measures important for robotics applications that leverage semantics are space complexity and learning speed~\cite{lesort2020cl_robotics}. We capture space complexity for each CKGE method using Model Size (MS) and Samples Storage Size (SSS) measures~\cite{lesort2020cl_robotics}. MS measures the growth in memory usage $\mathcal{U}$ for model parameters $\theta$ across learning sessions for a particular method---Equation~\ref{eq:ms}. Samples Storage Size (SSS) measures the growth in memory usage $\mathcal{U}$ for stored samples $SS$ across learning sessions as a proportion of the total number of training samples for the task, i.e., $\mathcal{D}_{Tr}$, in Equation~\ref{eq:sss}. For learning speed, we use the Learning Curve Area (LCA) measure~\cite{lesort2020cl_robotics}, which we modify to range between zero and one (like other measures). For a performance measure $m$, it computes the area covered by the learning curve of the learning method up to the best measured performance $m^{*}$ at time $t$ as a proportion of the area achieved by perfect zero-shot learning (Equation~\ref{eq:lca}).
\vspace{-0.3em}
\begin{equation}
    \small \textrm{\small MS}\!=\!\textrm{min}(1,\! \frac{\sum_{.  E}^{N}\frac{\mathcal{U}(\theta_{1})}{\mathcal{U}(\theta_{i})}}{N})
    \vspace{-0.5em}
    \label{eq:ms}
\end{equation}
\vspace{-0.5em}
\begin{equation}
    \small \textrm{\small SSS}\!=\!1\!-\!\textrm{min}(1,\!
    \frac{\sum_{i=1}^{N}\frac{\mathcal{U}(SS_{i})}{\mathcal{U}(\mathcal{D}_{Tr})}}{N})
    \vspace{-0.5em}
    \label{eq:sss}
\end{equation}
\vspace{-0.5em}
\begin{equation}
    \small \textrm{\small LCA}\!=\!\frac{\int_{0}^{t}m\,dm}{m^{*} \times t}
    \label{eq:lca}
\end{equation}

\textbf{Software implementation:} Please see supplementary material\footnote{\url{https://github.com/adaruna3/continual-kge}} for details about the tuning of hyper-parameters of CKGE methods, each knowledge graph embedding representation used for evaluation, and evaluation datasets, experiments, and results that are omitted here for brevity.

\begin{figure}[b!]
    \vspace{-2em}
    \centering
        \begin{subfigure}[b]{0.4\textwidth}
            \centering
            \includegraphics[width=1\linewidth]{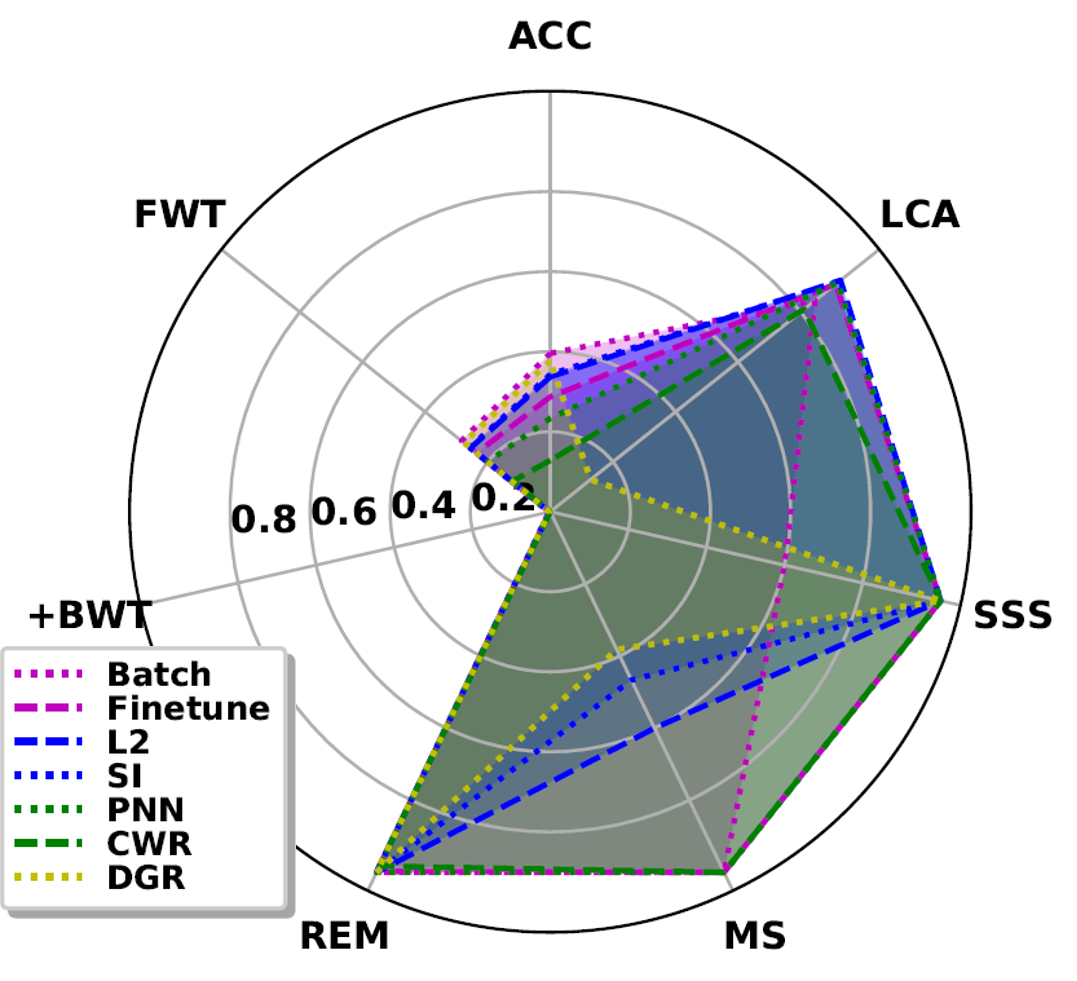}
            \caption{}
        \end{subfigure}
        \begin{subfigure}[b]{0.4\textwidth}
            \includegraphics[width=1\linewidth]{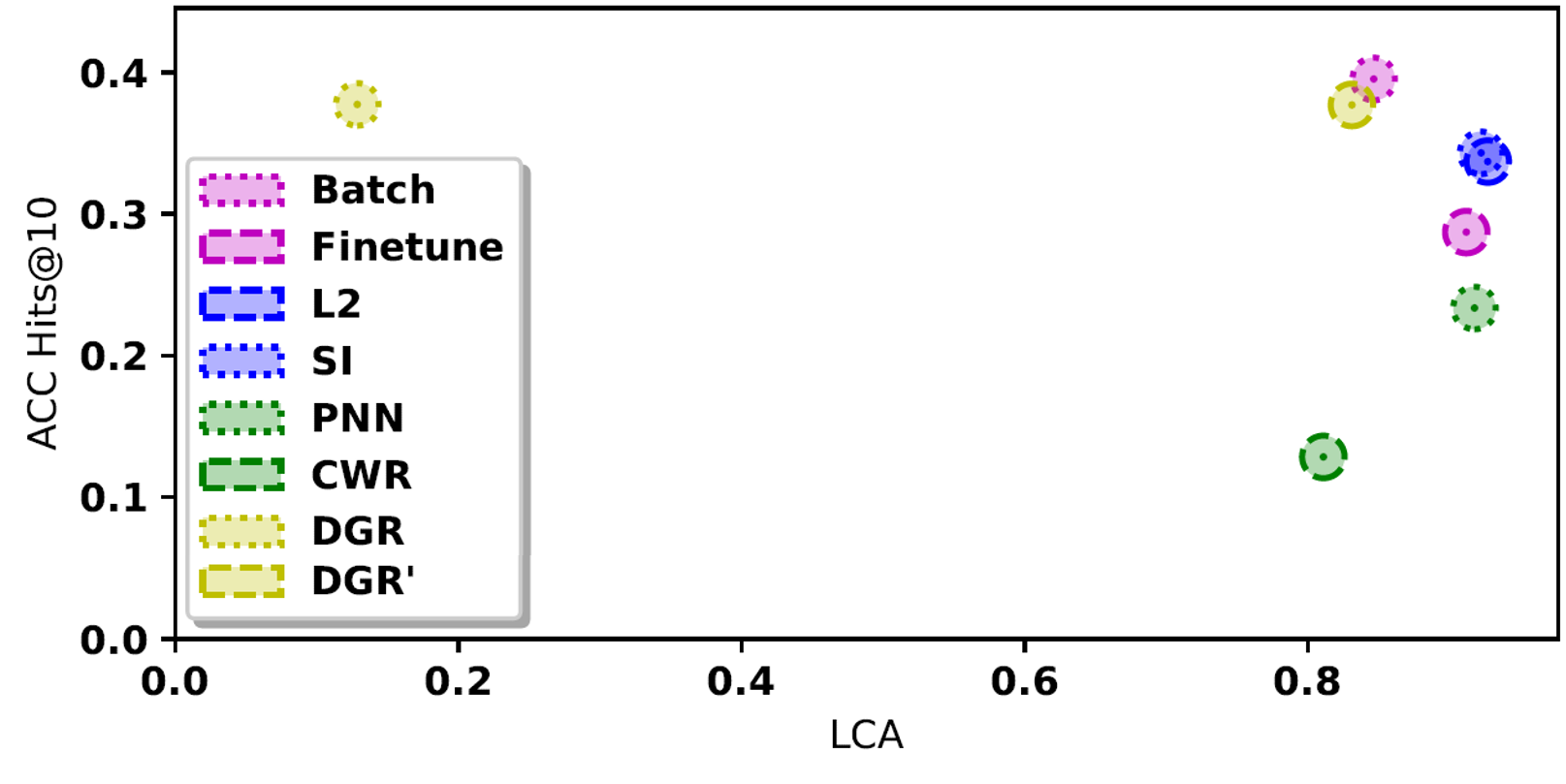}
            \caption{}
        \end{subfigure}
        \begin{subfigure}[b]{0.4\textwidth}
            \includegraphics[width=1\linewidth]{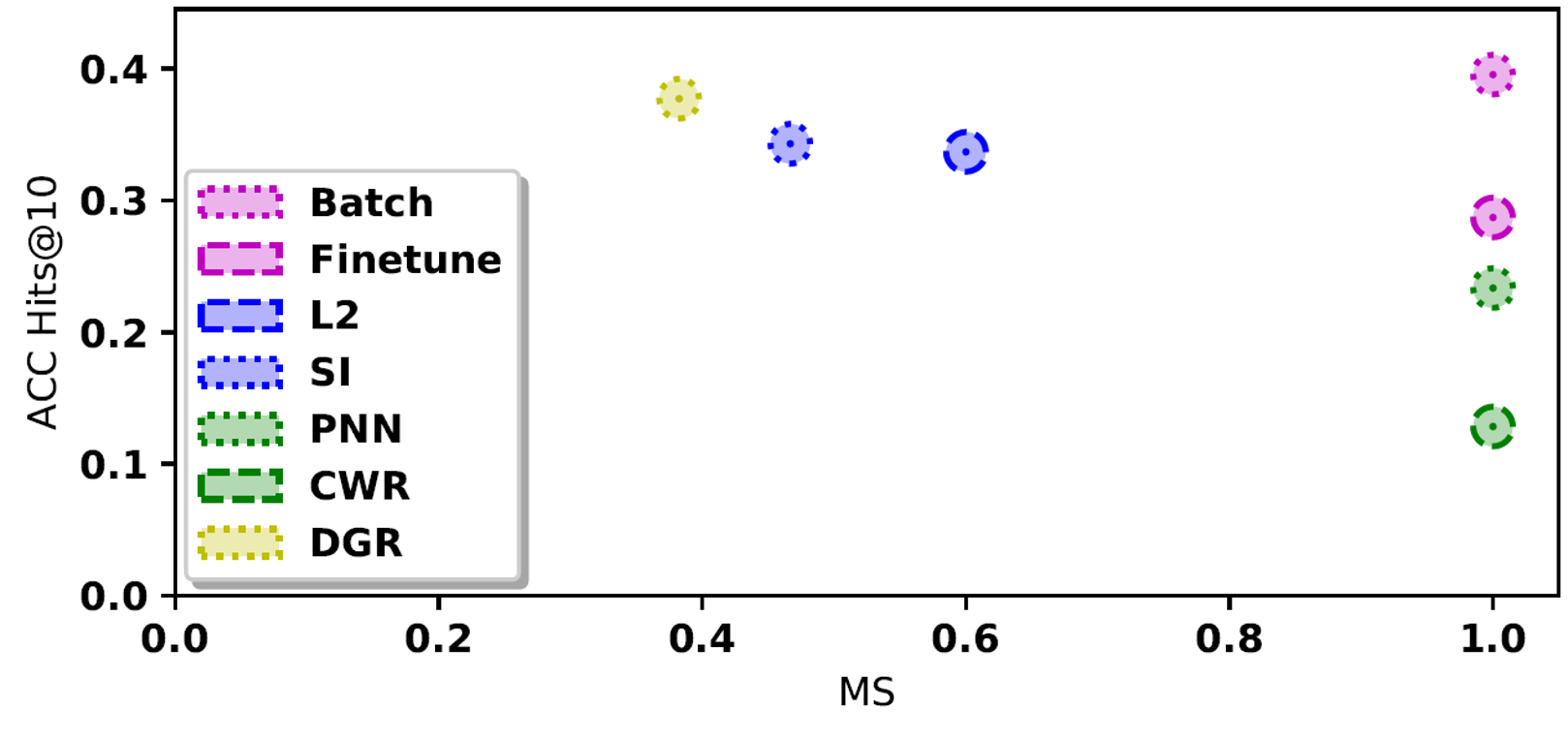}
            \caption{}
        \end{subfigure}
    \caption{\small Measures averaged for all datasets in Table~\ref{tbl:benchmark_datasets} and graph embedding representations in Section~\ref{sec:exp_pro}. Hits@10 used for ACC, FWT, +BWT, and REM. Best viewed in color.}
    \label{fig:bench_summary}
\end{figure}

\section{Experimental Results}
\label{sec:result}
Results reported in this section are the average of five test runs in each experimental scenario; statistical significance is tested using repeated-measures ANOVA and a post-hoc Tukey’s test. Any mention of ‘significance’ implies statistical significance at $95\%$ significance level (i.e. $p < 0.05$).

In addition to the CKGE methods, we considered two additional methods that served as upper and lower bounds (i.e., baselines) for the expected inference performance of the CKGE methods. \textit{Batch} represents the inference upper bound because it can store all prior examples to train a new embedding in each learning session. \textit{Finetune} represents the lower bound because it fine-tunes the embedding with examples only from the current learning session and has no means to prevent catastrophic forgetting.

\textbf{Benchmark evaluations:} Figure~\ref{fig:bench_summary}a summarizes the results of experiments using benchmark knowledge graph datasets of Table~\ref{tbl:benchmark_datasets} (WN18RR, FB15K237), where the range of each measure is $[0, 1]$ and larger values are better. Although DGR significantly outperforms other methods in terms of inference (i.e., using ACC and FWT), there are insights and trade-offs to consider based on other factors.

\begin{figure}[b!]
    \vspace{-0.5cm}
    \centering
    \includegraphics[width=0.9\columnwidth]{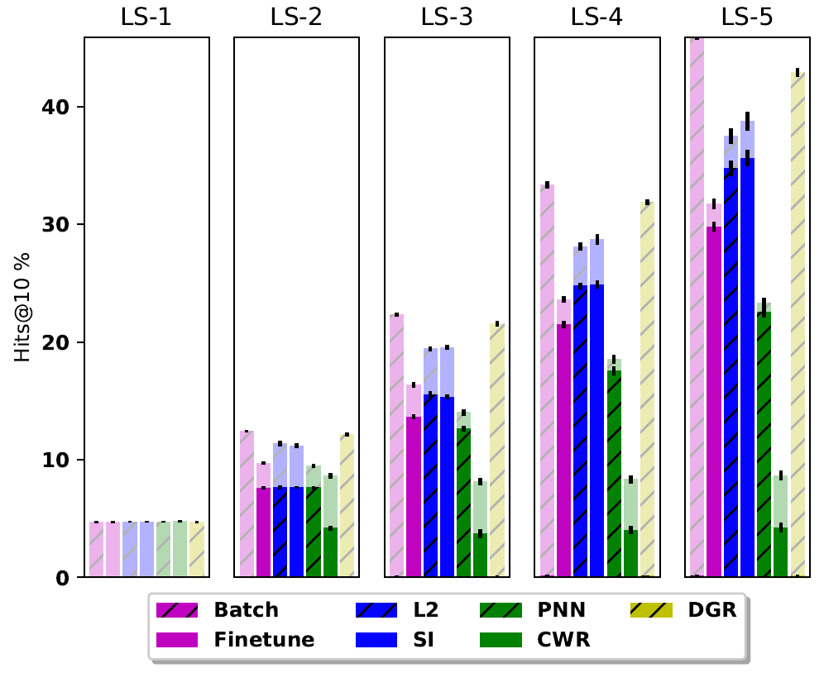}
	\caption[]
	{\small Hits@10 from initial (bright) to final (transparent) epoch. Black errors bars indicate standard deviation. L2R and SI perform better than DGR in the initial epoch, but DGR outperforms in the final epoch after the first learning session. Best viewed in color.}
	\label{fig:intialvsfinal}
\end{figure}

\begin{figure*}[h]
    \vspace{0.1cm}
    \centering
    \begin{tabular}{ccc}
      \raisebox{-4mm}{\includegraphics[width=54mm]{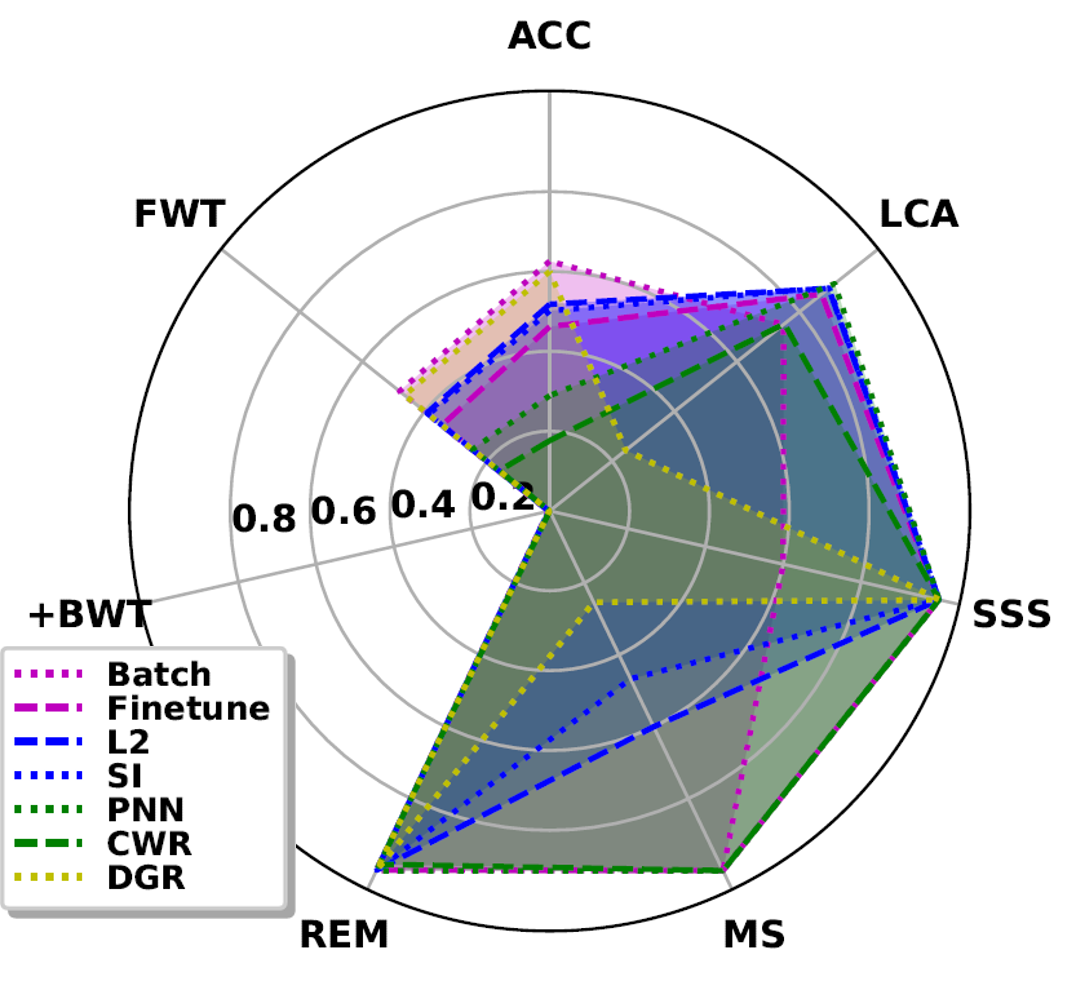}} & 
      \raisebox{-4mm}{\includegraphics[width=54mm]{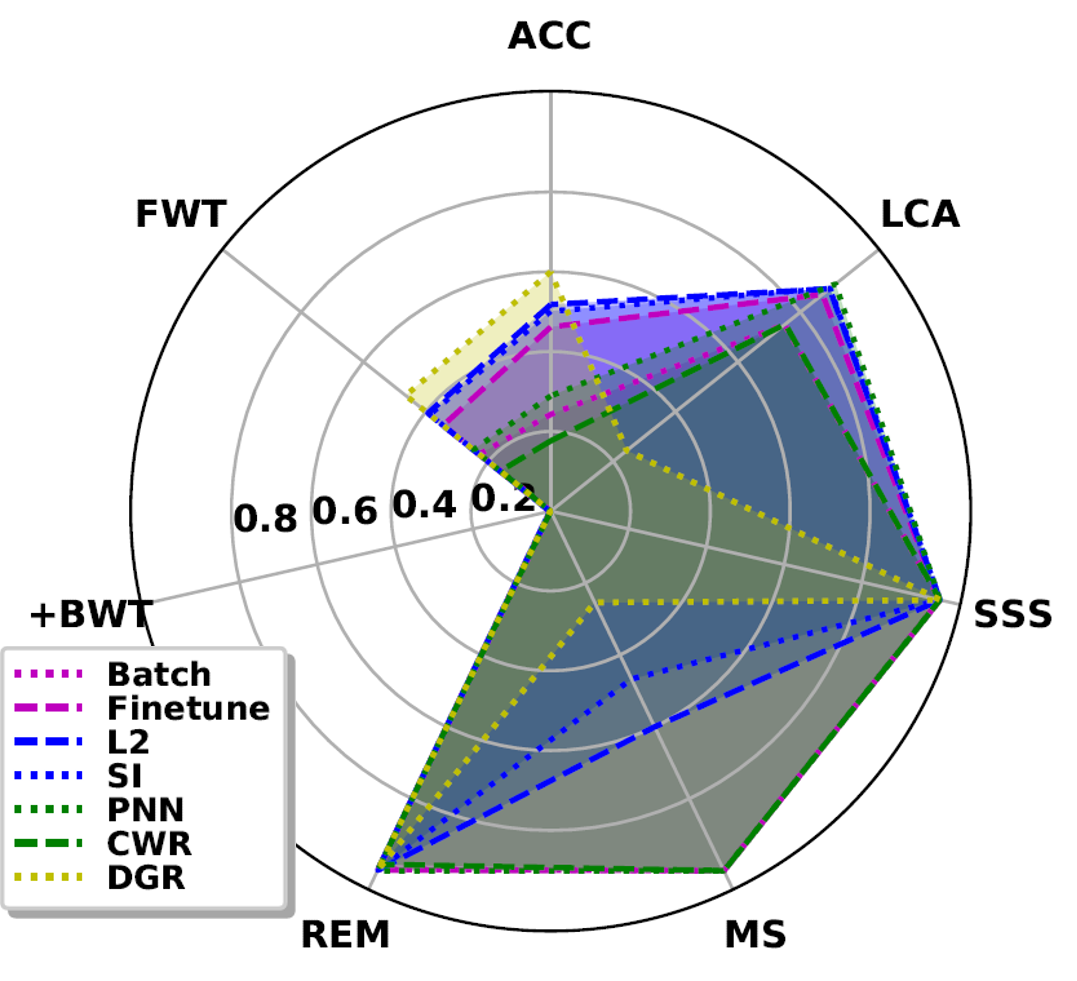}} & 
      \raisebox{-4mm}{\includegraphics[width=54mm]{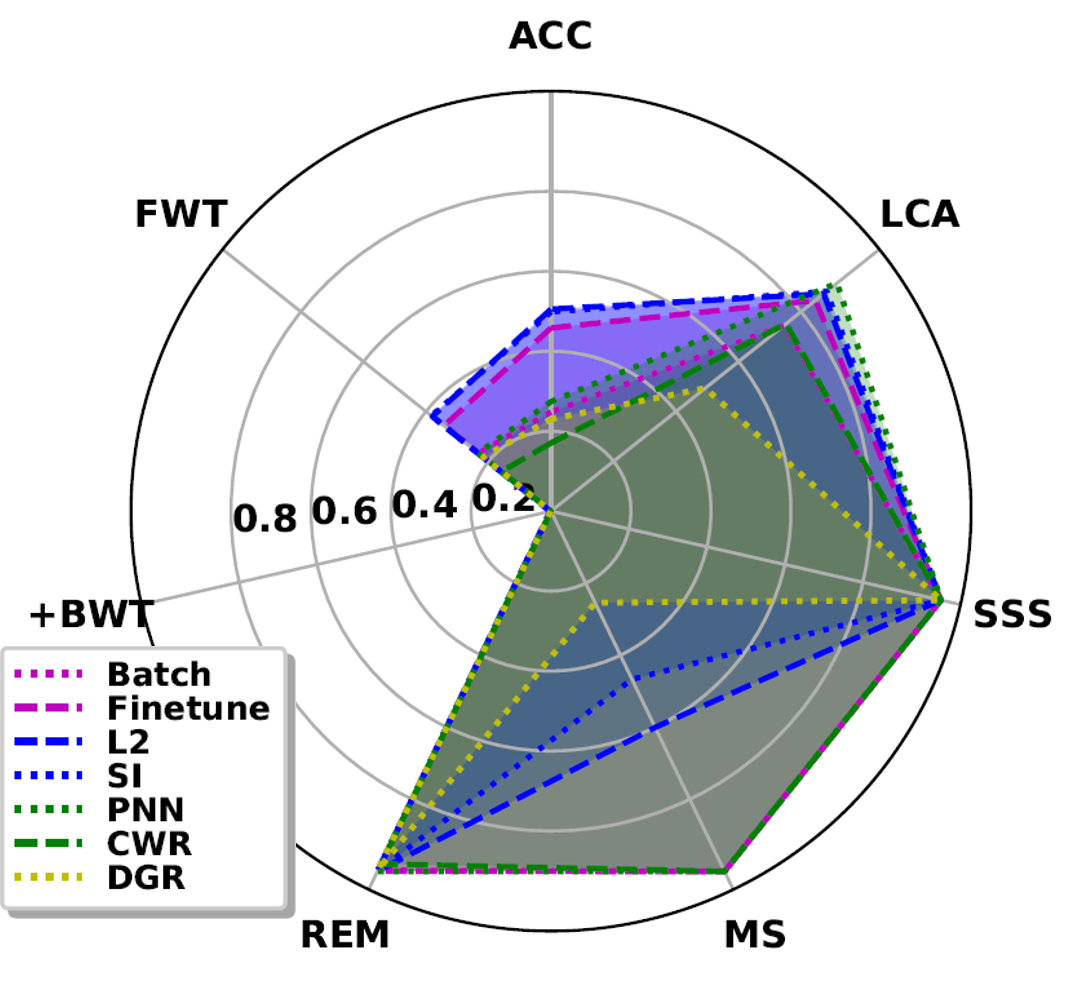}} \\
      \includegraphics[width=54mm]{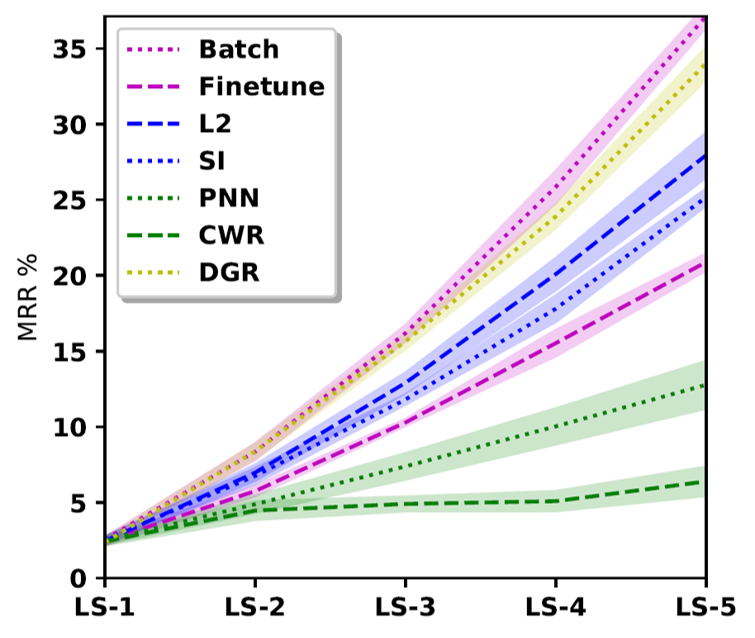} & 
      \includegraphics[width=54mm]{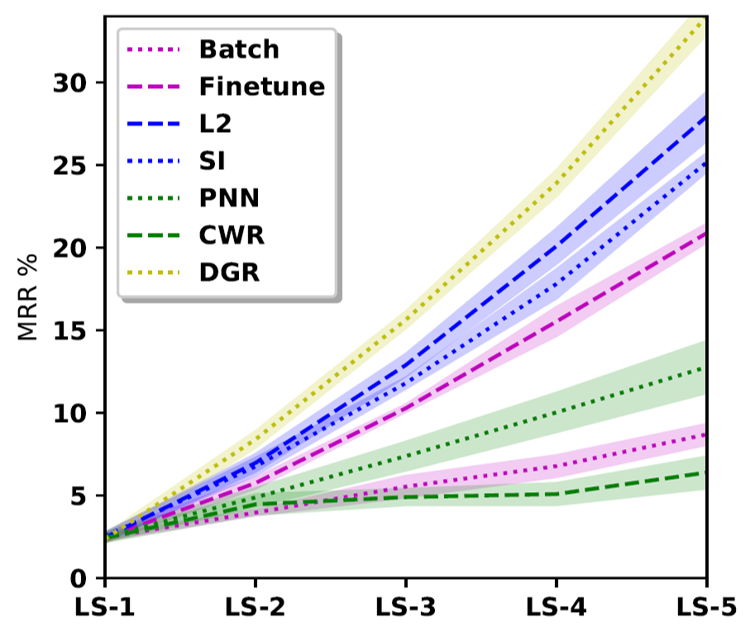} & 
      \includegraphics[width=54mm]{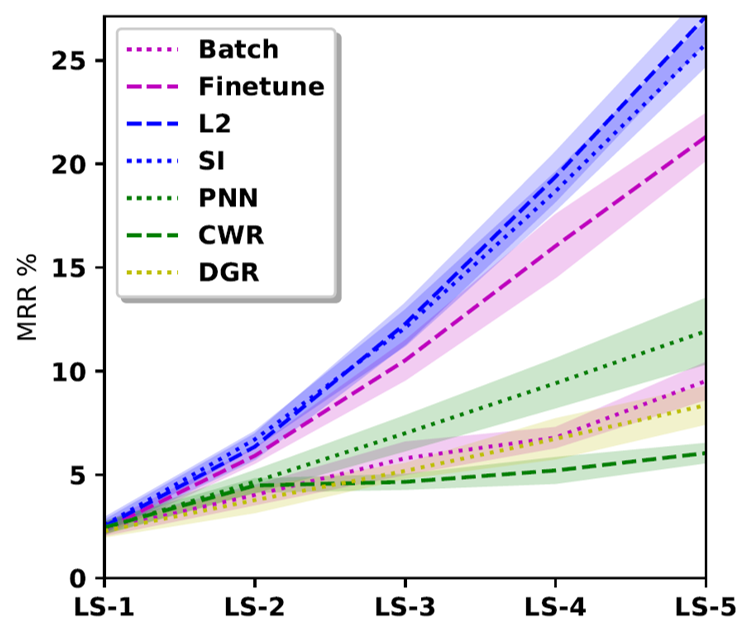} \vspace{-0.1cm} \\
      \small (a) & \small (b) & \small (c) \\[6pt]
    \end{tabular}
    \vspace{-1em}
    \caption{\small Semantics-driven robotics application scenarios: (a) Unconstrained; (b) Data Constrained; and (c) Time and Data Constrained. In each line plot, shading indicates standard deviation. Best viewed in color.}
    \vspace{-1.5em}
    \label{fig:robot_sum}
\end{figure*}

\begin{small_ind_s_itemize}
    \item Figure~\ref{fig:bench_summary}b shows that DGR has a significantly lower learning speed (based on LCA) than the other methods since a new generative model must be trained in each learning session. If the number of epochs to train the generative model are ignored, DGR's LCA is comparable to Batch (DGR$^{\prime}$ in Figure~\ref{fig:bench_summary}b) but still significantly lower than the regularization techniques (L2R and SI).
    
    \item Figure~\ref{fig:bench_summary}c indicates that methods with good inference performance also tend to have higher model memory growth (i.e., MS measure); among the methods with significantly better inference performance than Finetune, L2R has the smallest MS followed by SI and DGR.
    
    \item Since they regularize prior embeddings, L2R and SI initially perform better than DGR, as does the Finetune baseline, as seen in the Hits@10 plots for each method at the start and end of each learning session (Figure~\ref{fig:intialvsfinal}).
    
    \item Figures~\ref{fig:bench_summary} and~\ref{fig:intialvsfinal} indicate that the CKGE methods based on architecture modification (i.e., PNN and CWR) have significantly lower inference performance than Finetune in all experiments. The difference in performance between PNN and the regularization-based methods shows the importance of flexibility over prior concepts for CKGE. Also, CWR's poor inference performance highlights the challenges of directly manipulating the embedding space because, although CWR can learn TE well in isolation, CE is quickly corrupted by the averaging performed to merge embeddings.
\end{small_ind_s_itemize}

\textbf{Service robotics evaluation:} We constructed three evaluation scenarios using the AI2Thor knowledge graph dataset in Table~\ref{tbl:robot_dataset}. Each scenario corresponds to a different class of semantics-driven robotics applications. The first scenario, \textit{Unconstrained} in Figure~\ref{fig:robot_sum}a, corresponds to a robot that has access to all prior training examples at training time. More generally, this could represent robots with ready access to cloud services for data storage and processing. However, such a scenario may be unfeasible in some applications due to hardware constraints or security concerns. Our second scenario, \textit{Data Constrained} in Figure~\ref{fig:robot_sum}b, represents robots with access to limited training examples, e.g., only from the current learning session ($\mathcal{D}^{n}_{Tr}$); this could be due to storage constraints or dynamic domain changes. The final scenario, \textit{Time and Data Constrained} in Figure~\ref{fig:robot_sum}c, mimics a mobile robot (or drone) operating under resource constraints; the robot only has access to training examples for the current learning session and has limited time to update the knowledge graph embedding. For simplicity, we limited the number of training epochs in each learning session to 100. The ranges of each measure are in $[0,1]$ and larger values are better. The results from each scenario provide key insights about the choice of the CKGE method:
\begin{small_ind_s_itemize}
    \item In an \textit{unconstrained scenario} (Figure~\ref{fig:robot_sum}a), such as one in which a robot might have access to a cloud compute service, Batch learning is the best choice despite its significantly lower sample efficiency (SSS) and learning speed (LCA) because it provides significantly higher ACC and FWT compared with other methods.
    
    \item In a \textit{data constrained scenario} (Figure~\ref{fig:robot_sum}b), e.g., the robot can only update its semantic representation intermittently using limited on-board hardware. Batch's inference performance collapses because prior observations are unavailable. Given these constraints, DGR is the best choice, with much better ACC and FWT than other methods because it approximates Batch in the unconstrained scenario. However, DGR incurs a significant computational cost to train the generative model, resulting in a significantly lower LCA value.
    
    \item In a \textit{data and time constrained scenario} (Figure~\ref{fig:robot_sum}c), e.g., the robot is updating its own semantic model on-board \textit{during} a task, DGR is a poor choice because there is not enough time to sufficiently train the generative model. L2R and SI are better choices; SI with Analogy and L2R with either graph embedding offer significantly better inference performance than Finetune and significantly better LCA than Batch. Compared with SI, L2R's memory growth (MS) is significantly lower.
\end{small_ind_s_itemize}

\section{Conclusion}
\label{sec:conclusion}
Knowledge graph embeddings are increasingly being used as semantic representations in robotics applications, but it is difficult to update these representations incrementally. This paper introduced five representative continual learning-inspired methods for continual knowledge graph embedding (CKGE). We also introduced a heuristic sampling strategy and generated CKGE datasets based on benchmark knowledge graphs and a knowledge graph for the service robotics domain. Furthermore, we identified and built on measures for evaluating continual learning in robotics. We evaluated our embedding-generic methods on two knowledge graph embedding representations. Experimental evaluation using the benchmark knowledge graphs provided key insights characterizing the use of our CKGE methods in terms of factors such as inference, learning speed, and memory requirements. Our evaluation using the service robotics domain knowledge characterized the use of CKGE methods in three different classes of semantics-driven robotics applications. Future work will further investigate the adaptation of continual learning principles for CKGE in robot tasks that require semantic knowledge representations, including data from physical robots in complex, dynamic domains.


\bibliographystyle{./IEEEtran}
\balance{\bibliography{references}}  

\newpage  
\appendix 
In addition to our sampling strategy in Section~\ref{sec:exp_pro}, referred to here as {\it triple} sampling, we also experimented with entity and relation sampling. We experimented with triple, relation, and entity sampling strategies to get a breadth of sampling strategies one could use to generate a CKGE dataset from an established benchmark knowledge graph. For our main results, we chose triple sampling because it had the most fidelity to the original benchmark knowledge graphs, while still modeling a challenging sampling scenario a robot might encounter. By providing results from experiments with all sampling strategies, we show that the insights for our main results hold for different sampling strategies and that our main narrative avoids highlighting outlier results. Below we detail how the entity and relation sampling strategies were implemented, the CKGE datasets generated by each, and results obtained for these datasets using experimental settings similar to Section~\ref{sec:result}.

\textbf{CKGE sampling strategies:} Our entity sampling strategy closely follows the sampling strategy proposed in \cite{song2018enriching}. As in Section~\ref{sec:exp_pro}, consider a knowledge graph $\mathcal{G}$ whose triples $\mathcal{D}$ have been split into a training set $\mathcal{D}_{Tr}$, validation set $\mathcal{D}_{Va}$, and test set $\mathcal{D}_{Te}$. Our approach for generating entity sampling datasets for $n=\{1,...,N\}$ learning sessions is:
\begin{small_ind_enumerate}
    \item \textit{Initialize entity sampling distribution}: initialize the entity sampling distribution to uniform likelihood for any entity to be sampled.
    
    \item \textit{Sample entities}: sample without replacement $50\%$ of entities in $\mathcal{E}$ of $\mathcal{G}$.
   
    \item \textit{Extract relations}: create a set of entities $E^{n}$ and a set of relations $R^{n}$ for this session from the sampled entities and the relations of triples in $\mathcal{G}$ that connect any two entities in $E^{n}$, respectively. The set of all observed entities (relations), i.e., $\mathcal{E}^{n}$ ($\mathcal{R}^{n}$) is the union of current and prior $E^{n}$ ($R^{n}$).
   
    \item \textit{Construct $n^{th}$ train, validation, and test sets}: extract from $\mathcal{D}_{Tr}$, $\mathcal{D}_{Va}$, and $\mathcal{D}_{Te}$ the triples whose head, relation, and tail belong to $E^{n}$ and $R^{n}$ (respectively). These triples form train set $\mathcal{D}^{n}_{Tr}$, validation set $\mathcal{D}^{n}_{Va}$ and test set $\mathcal{D}^{n}_{Te}$ of the $n^{th}$ session.
   
    \item \textit{Update entity distribution and repeat}: to bias towards sampling new triples, decrease the likelihoods of sampled entities in $E^{n}$ to be sampled in future learning sessions in proportion to the number of times an entity has been sampled.
    
    \item Repeat steps 2-5 until a predefined number of iterations are completed.
\end{small_ind_enumerate}

We generated two CKGE datasets with $n=5$ sessions using the entity sampling approach on two established benchmark knowledge graphs in the graph embedding community (WN18RR and FB15K237~\cite{dettmers2018convolutional}). Table~\ref{tbl:entity_datasets} reports statistics of each dataset as in Section~\ref{sec:exp_pro}. Note that even after 5 learning sessions, not all training, validation, nor test triples have been sampled from the benchmark datasets. Additionally, the number of triples in each learning session exceeds that of triple sampling, proving how this method often repeats sampled triples across learning sessions.

\begin{table}[h!]
    \setlength{\tabcolsep}{4pt} 
    \vspace{-0.1cm}
    \caption{\centering \small CKGE Datasets; Entity Sampling Benchmarks}
	\vspace{-0.2cm}
    \resizebox{\columnwidth}{!}{
        \begin{tabular}{l|ccccc} \toprule
        & \multicolumn{5}{c}{WN18RR-5-LS} \\ \midrule
        & LS-1 & LS-2 & LS-3 & LS-4 & LS-5 \\ \midrule
        $|E^{n}|$ & 20,471/(50\%) & 20,471/(81\%) & 20,471/(95\%) & 20,471/(99\%) & 20,471/(100\%) \\ 
        $|R^{n}|$ & 11/(100\%) & 11/(100\%) & 11/(100\%) & 11/(100\%) & 11/(100\%) \\ 
        $|\mathcal{D}^{n}_{Tr}|$ & 21,729/(25\%) & 21,937/(47\%) & 21,888/(63\%) & 20,870/(74\%) & 21,852/(83\%) \\ 
        $|\mathcal{D}^{n}_{Va}|$ & 773/(25\%) & 774/(48\%) & 796/(64\%) & 727/(75\%) & 801/(84\%) \\ 
        $|\mathcal{D}^{n}_{Te}|$ & 821/(26\%) & 800/(48\%) & 798/(64\%) & 729/(75\%) & 797/(83\%) \\ \midrule
        & \multicolumn{5}{|c}{FB15K237-5-LS} \\ \midrule
        & LS-1 & LS-2 & LS-3 & LS-4 & LS-5 \\ \midrule
        $|E^{n}|$ & 7,270/(50\%) & 7,270/(81\%) & 7,270/(95\%) & 7,270/(99\%) & 7,270/(100\%) \\
        $|R^{n}|$ & 230/(97\%) & 231/(99\%) & 226/(100\%) & 231/(100\%) & 230/(100\%) \\ 
        $|\mathcal{D}^{n}_{Tr}|$ & 73,846/(27\%) & 69,288/(48\%) & 66,326/(64\%) & 70,519/(75\%) & 73,565/(84\%) \\ 
        $|\mathcal{D}^{n}_{Va}|$ & 4,683/(27\%) & 4,616/(49\%) & 4,365/(65\%) & 4,411/(76\%) & 4,932/(84\%) \\ 
        $|\mathcal{D}^{n}_{Te}|$ & 5,569/(27\%) & 5,427/(49\%) & 4,997/(65\%) & 5,218/(76\%) & 5,774/(84\%) \\ \bottomrule
        \end{tabular}
        \label{tbl:entity_datasets}
    }
    \vspace{0.1cm}
	\caption{\centering \small CKGE Datasets; Relation Sampling Benchmarks}
    \resizebox{\columnwidth}{!}{
        \begin{tabular}{l|ccccc} \toprule
        & \multicolumn{5}{c}{WN18RR-5-LS} \\ \midrule
        & LS-1 & LS-2 & LS-3 & LS-4 & LS-5 \\ \midrule
        $|E^{n}|$ & 16,564/(40\%) & 12,447/(47\%) & 35,670/(96\%) & 37,548/(99\%) & 14,062/(99\%) \\
        $|R^{n}|$ & 5/(45\%) & 5/(82\%) & 5/(91\%) & 5/(100\%) & 5/(100\%) \\ 
        $|\mathcal{D}^{n}_{Tr}|$ & 16,886/(19\%) & 12,840/(26\%) & 37,566/(66\%) & 69,555/(100\%) & 14,442/(100\%) \\ 
        $|\mathcal{D}^{n}_{Va}|$ & 216/(7\%) & 201/(11\%) & 490/(25\%) & 1,979/(75\%) & 184/(76\%) \\ 
        $|\mathcal{D}^{n}_{Te}|$ & 205/(7\%) & 228/(11\%) & 511/(26\%) & 2053/(75\%) & 185/(76\%) \\ \midrule
        & \multicolumn{5}{|c}{FB15K237-5-LS} \\ \midrule
        & LS-1 & LS-2 & LS-3 & LS-4 & LS-5 \\ \midrule
        $|E^{n}|$ & 13,666/(94\%) & 13,522/(98\%) & 13,704/(100\%) & 13,023/(100\%) & 13,989/(100\%) \\ 
        $|R^{n}|$ & 118/(50\%) & 118/(80\%) & 118/(95\%) & 118/(99\%) & 118/(100\%) \\ 
        $|\mathcal{D}^{n}_{Tr}|$ & 128,045/(47\%) & 148,446/(78\%) & 154,624/(98\%) & 113,904/(100\%) & 155,556/(100\%) \\ 
        $|\mathcal{D}^{n}_{Va}|$ & 7,546/(43\%) & 9,814/(78\%) & 9,474/(98\%) & 6,394/(99\%) & 11,184/(99\%) \\ 
        $|\mathcal{D}^{n}_{Te}|$ & 8,711/(43\%) & 11,378/(78\%) & 11,062/(98\%) & 7,559/(99\%) & 13,099/(99\%) \\ \bottomrule
        \end{tabular}
        \label{tbl:relation_datasets}
    }
\end{table}

Due to the large number of triples or entities sampled in the first learning session by both triple and entity sampling strategies, fewer than $5\%$ of all future relations sampled will be new, as shown across Tables~\ref{tbl:benchmark_datasets} and \ref{tbl:entity_datasets}. Therefore, the third and final sampling strategy we experimented with was relation sampling to encourage sampling new relations in later learning sessions. Again, consider a knowledge graph $\mathcal{G}$ whose triples $\mathcal{D}$ have been split into a training set $\mathcal{D}_{Tr}$, validation set $\mathcal{D}_{Va}$, and test set $\mathcal{D}_{Te}$. Our approach for generating relation sampling datasets for $n=\{1,...,N\}$ learning sessions is:
\begin{small_ind_enumerate}
    \item \textit{Initialize relation sampling distribution}: initialize the relation sampling distribution to uniform likelihood for any relation to be sampled.
    
    \item \textit{Sample relations}: sample without replacement $50\%$ of relations in $\mathcal{R}$ of $\mathcal{G}$.
   
    \item \textit{Extract entities}: create a set of relations $R^{n}$ and a set of entities $E^{n}$ for this session from the sampled relations and the entities of triples in $\mathcal{G}$ that are connected any relation in $R^{n}$, respectively. The set of all observed entities (relations), i.e., $\mathcal{E}^{n}$ ($\mathcal{R}^{n}$) is the union of current and prior $E^{n}$ ($R^{n}$).
   
    \item \textit{Construct $n^{th}$ train, validation, and test sets}: extract from $\mathcal{D}_{Tr}$, $\mathcal{D}_{Va}$, and $\mathcal{D}_{Te}$ the triples whose head, relation, and tail belong to $E^{n}$ and $R^{n}$ (respectively). These triples form train set $\mathcal{D}^{n}_{Tr}$, validation set $\mathcal{D}^{n}_{Va}$ and test set $\mathcal{D}^{n}_{Te}$ of the $n^{th}$ session.
   
    \item \textit{Update relation distribution and repeat}: to bias towards sampling new triples, decrease the likelihoods of sampled relations in $R^{n}$ to be sampled in future learning sessions in proportion to the number of times an relation has been sampled.
    
    \item Repeat steps 2-5 until a predefined number of iterations are completed.
\end{small_ind_enumerate}

We generated two CKGE datasets with $n=5$ sessions using the relation sampling approach on two established benchmark knowledge graphs in the graph embedding community (WN18RR and FB15K237~\cite{dettmers2018convolutional}). Table~\ref{tbl:relation_datasets} reports statistics of each dataset as in Section~\ref{sec:exp_pro}. While relation sampling uniquely has more new relations in later learning sessions, the strategy does not model a realistic world scenario a robot might face when exploring an environment. 

Our results obtained for these datasets support the results presented in Section~\ref{sec:result}, with some minor differences depending on the dataset or embedding method being used. Similar to Section~\ref{sec:result}, results reported below are the average of five test runs in each experimental scenario; statistical significance is tested using repeated-measures ANOVA and a post-hoc Tukey’s test. Any mention of ‘significance’ implies statistical significance at $95\%$ significance level (i.e. $p < 0.05$). As in Section~\ref{sec:result}, \textit{Batch} and \textit{Finetune} baselines are reported as upper and lower bounds for the expected inference performance of the CKGE methods.

\begin{figure}[t]
    \centering
        \begin{subfigure}[b]{0.4\textwidth}
            \centering
            \includegraphics[width=1\linewidth]{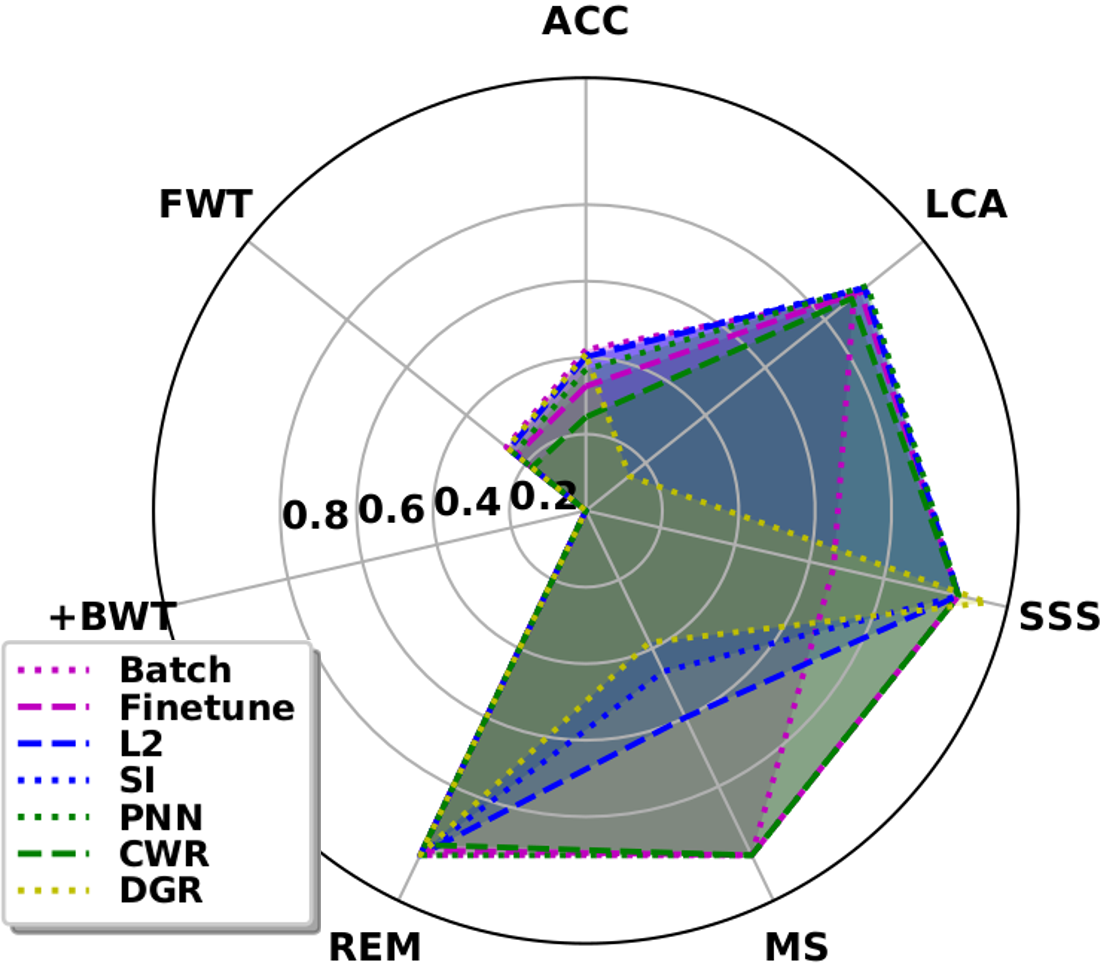}
            \caption{}
        \end{subfigure}
        \begin{subfigure}[b]{0.4\textwidth}
            \includegraphics[width=1\linewidth]{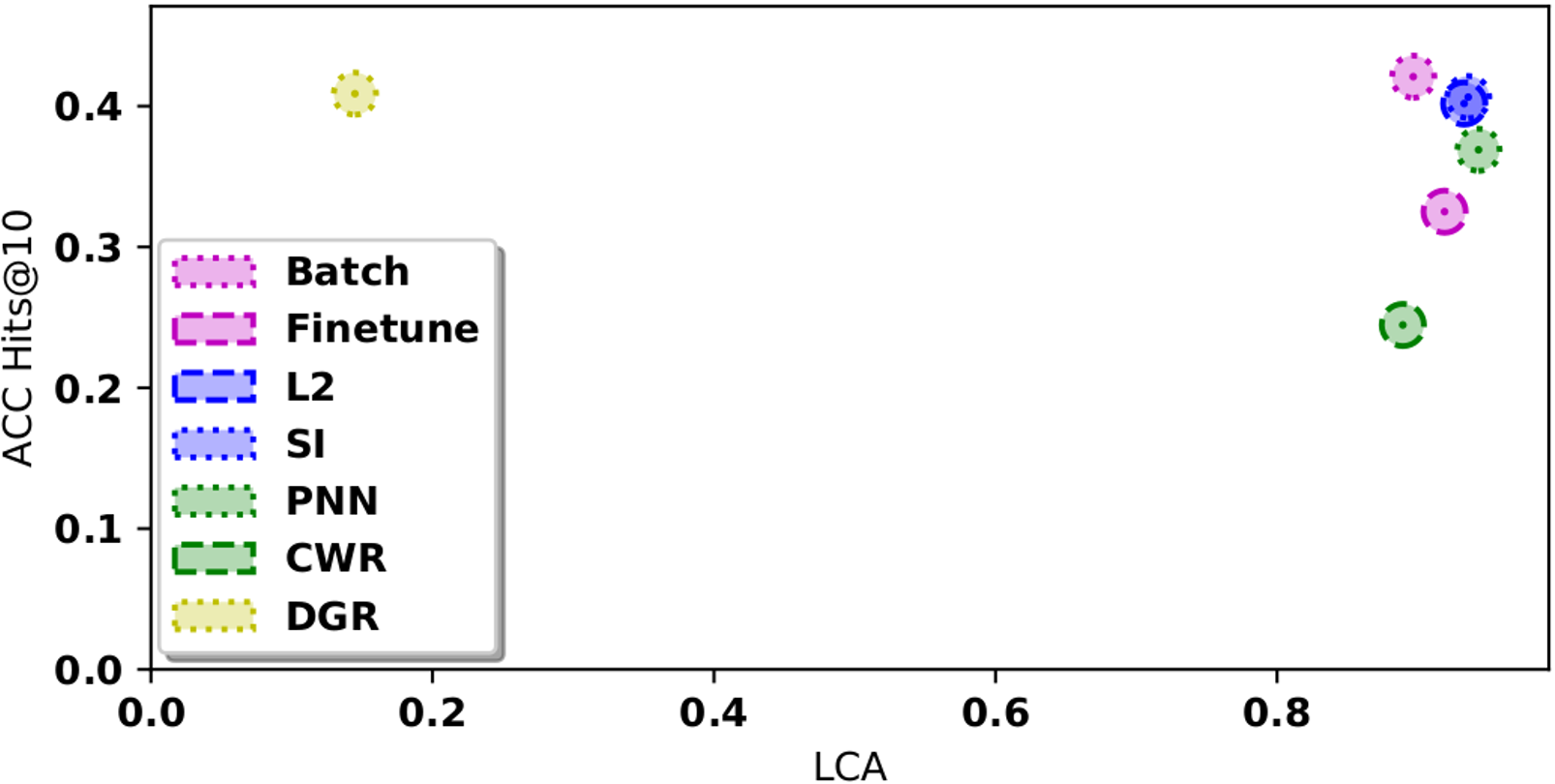}
            \caption{}
        \end{subfigure}
        \begin{subfigure}[b]{0.4\textwidth}
            \includegraphics[width=1\linewidth]{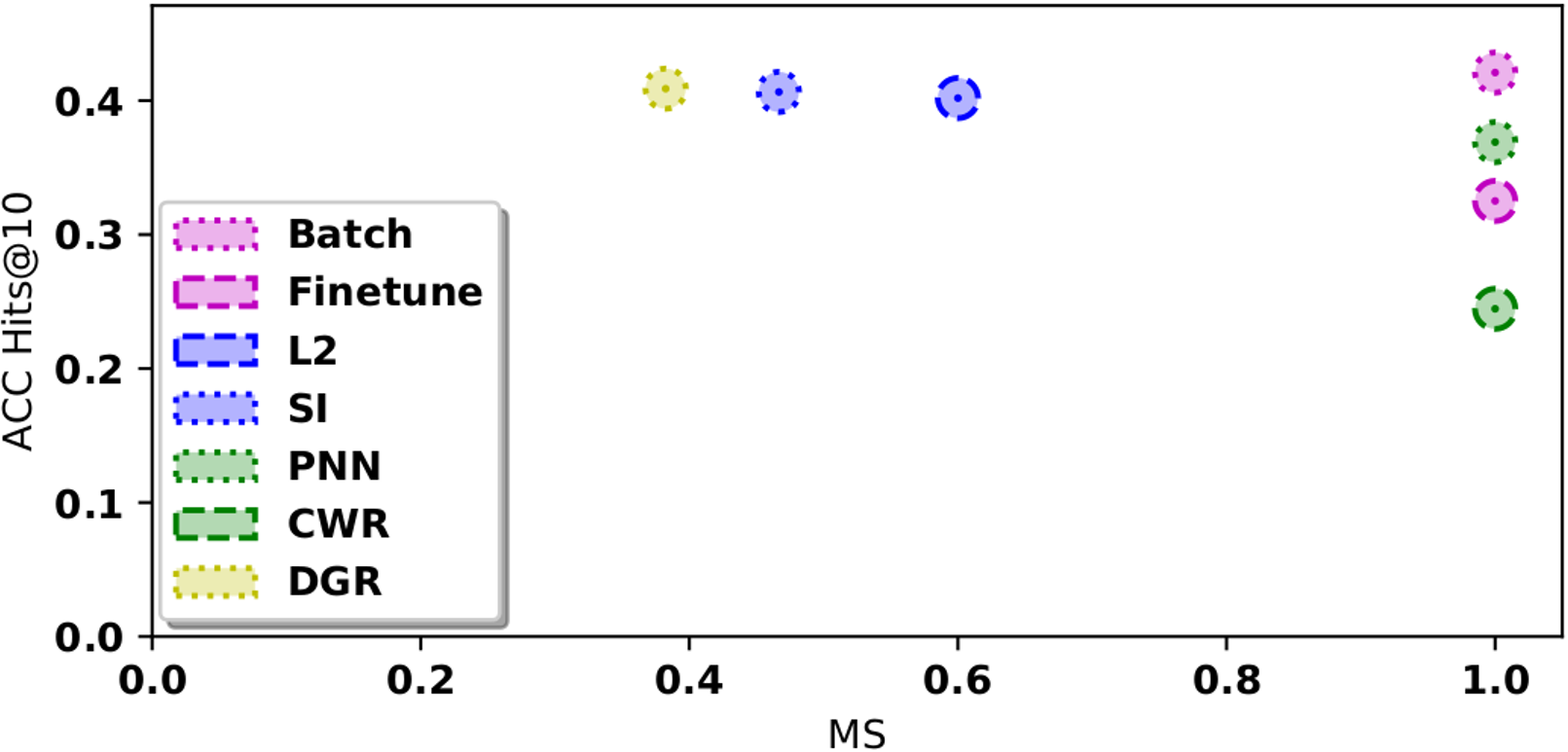}
            \caption{}
        \end{subfigure}
    \caption{\small Measures averaged for all datasets in Table~\ref{tbl:entity_datasets} and graph embedding representations in Section~\ref{sec:exp_pro}. Hits@10 used for ACC, FWT, +BWT, and REM. Best viewed in color.}
    \label{fig:entity_summary}
\end{figure}

\textbf{Entity sampling evaluations:} Figure~\ref{fig:entity_summary}a summarizes the results of experiments using entity sampling knowledge graph datasets of Table~\ref{tbl:entity_datasets} (WN18RR, FB15K237), where the range of each measure is $[0, 1]$ and larger values are better. Although DGR significantly outperforms other methods in terms of inference (i.e., using ACC), there are insights and trade-offs to consider based on other factors.

\begin{small_ind_s_itemize}
    \item Figure~\ref{fig:entity_summary}b shows that DGR has a significantly lower learning speed (based on LCA) than the other methods since a new generative model must be trained in each learning session. PNN had the best learning speed, having significantly better learning speeds than Batch across all datasets and embedding methods. Following close behind were the regularization techniques (L2R and SI), which had significantly better learning speeds than Batch in all cases except when using the Analogy embedding method for WN18RR.
    
    \item Figure~\ref{fig:entity_summary}c indicates that methods with good inference performance also tend to have higher model memory growth (i.e., MS measure); among the methods with significantly better inference performance than Finetune, PNN had the smallest MS followed by L2R, SI and DGR.
    
    \item PNN outperforms Finetune in this dataset shown in Figure~\ref{fig:entity_summary}. The difference in PNN performance from triple to entity sampling is likely because more entities are frozen sooner in the triple sampling datasets with fewer observed training triples than in the entity sampling datasets, as evident in the entity coverage statistics for each dataset.
    
    \item Figure~\ref{fig:entity_summary} indicates that the CWR has significantly lower inference performance than Finetune in all experiments. CWR's poor inference performance is again attributable to the challenges of directly manipulating the embedding space. 
\end{small_ind_s_itemize}

\begin{figure}[t]
    \centering
        \begin{subfigure}[b]{0.4\textwidth}
            \centering
            \includegraphics[width=1\linewidth]{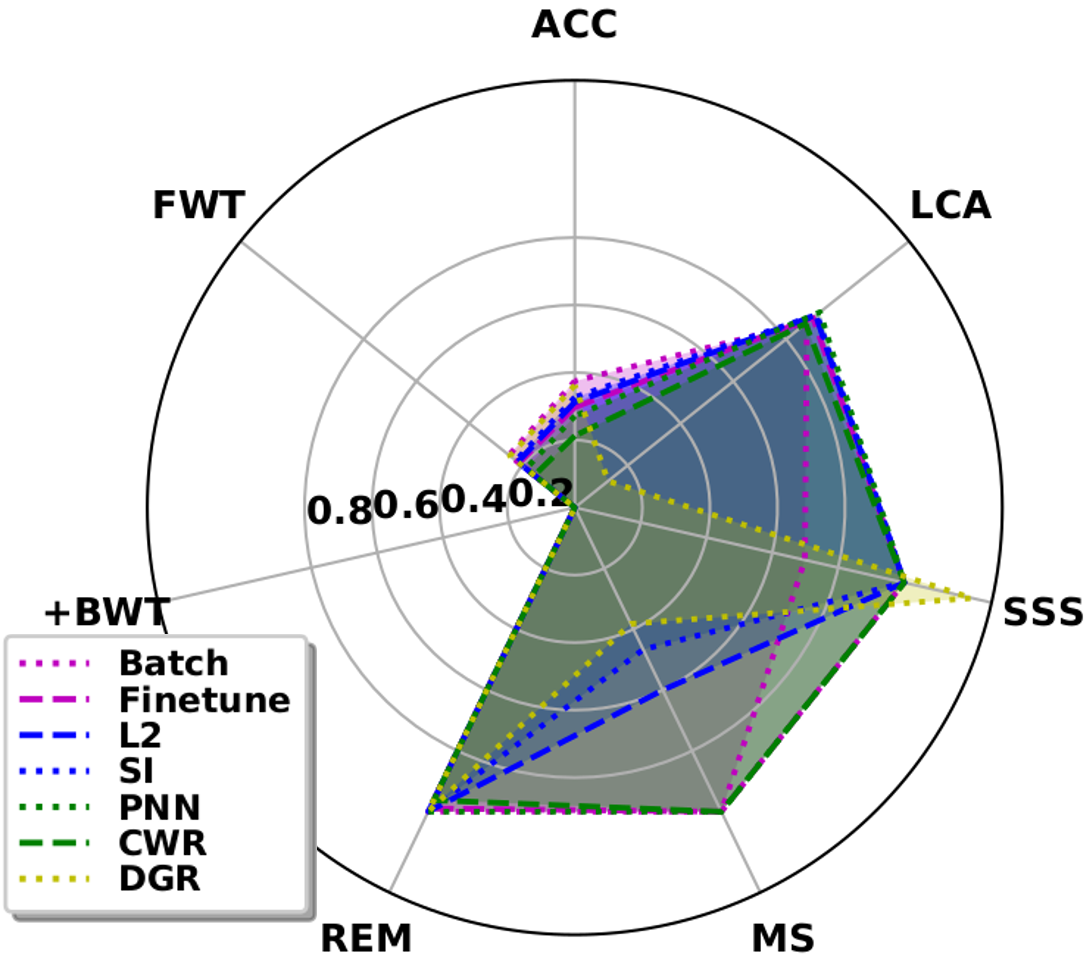}
            \caption{}
        \end{subfigure}
        \begin{subfigure}[b]{0.4\textwidth}
            \includegraphics[width=1\linewidth]{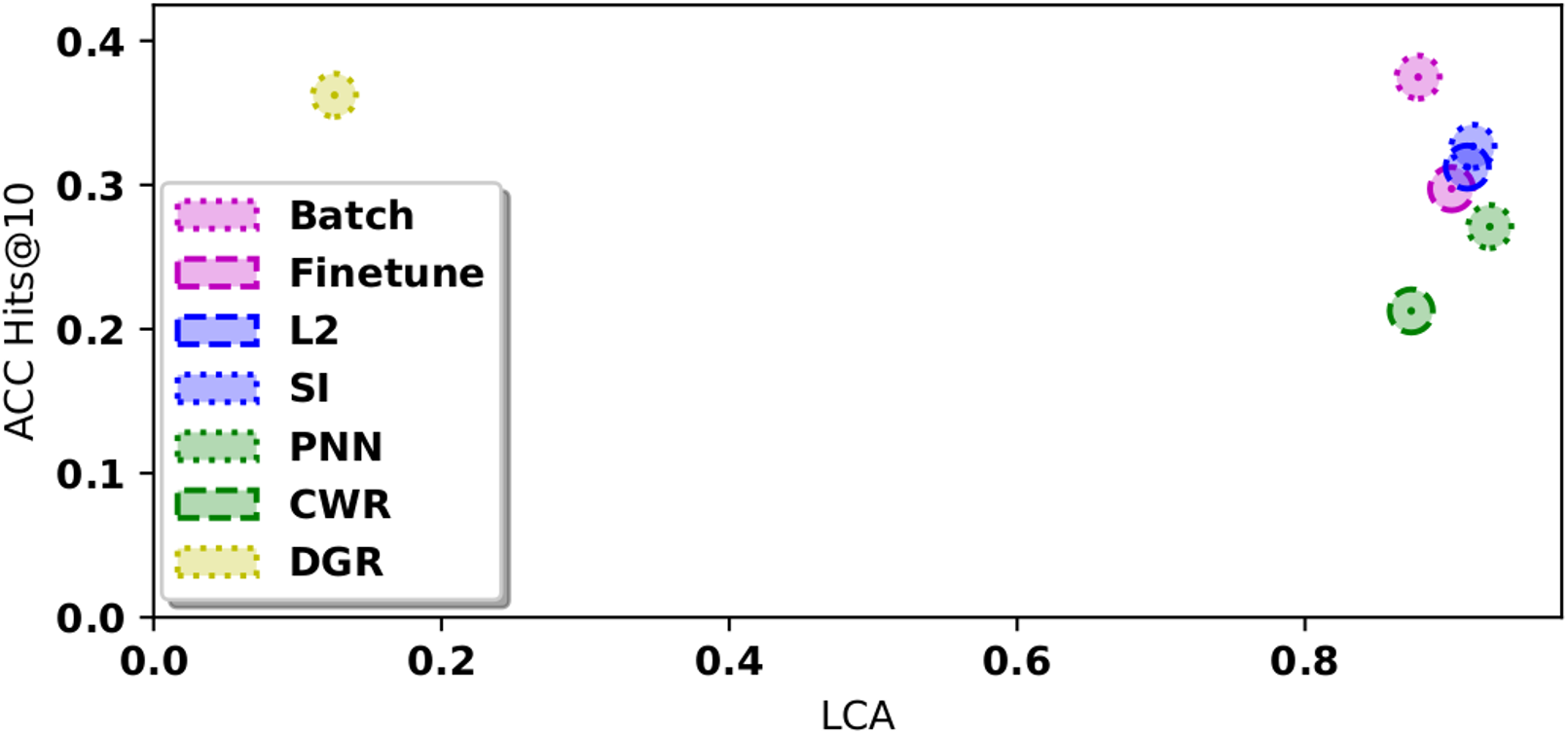}
            \caption{}
        \end{subfigure}
        \begin{subfigure}[b]{0.4\textwidth}
            \includegraphics[width=1\linewidth]{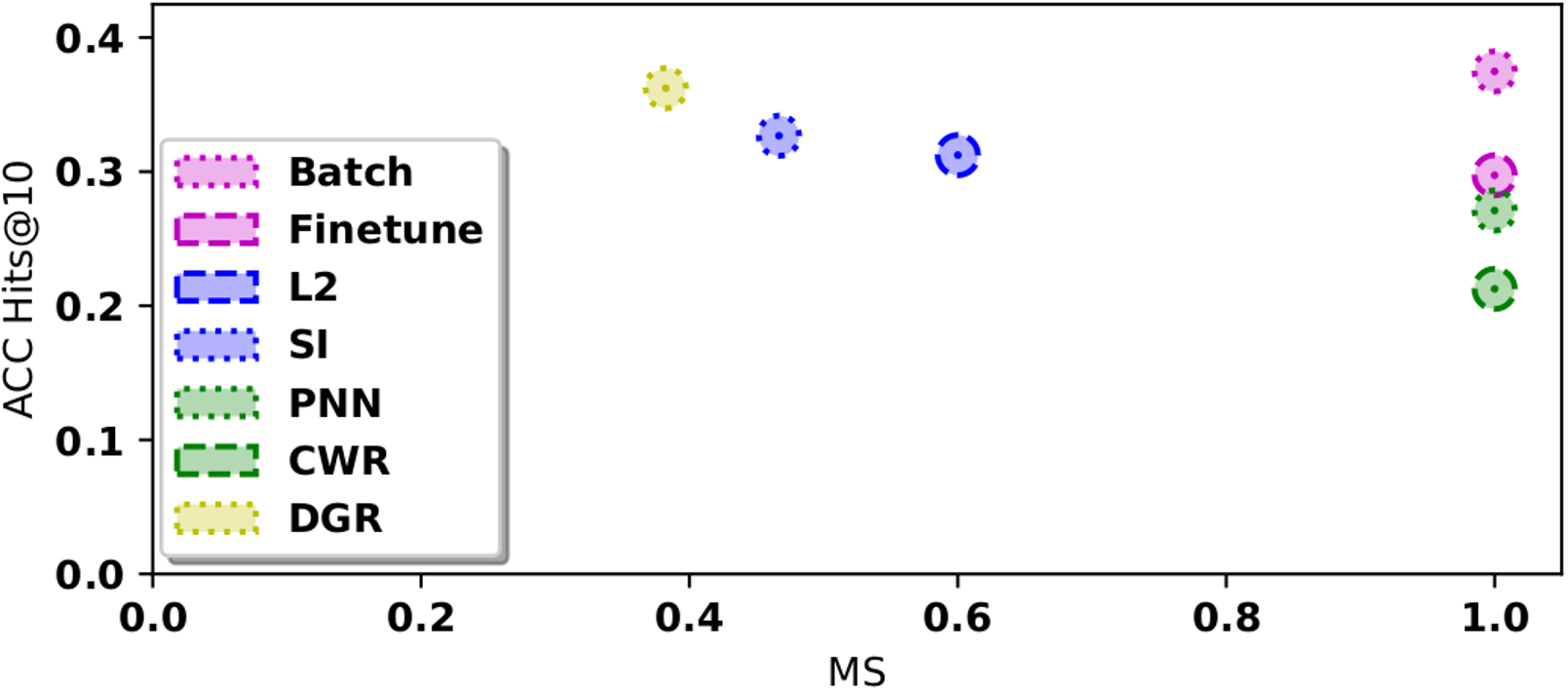}
            \caption{}
        \end{subfigure}
    \caption{\small Measures averaged for all datasets in Table~\ref{tbl:relation_datasets} and graph embedding representations in Section~\ref{sec:exp_pro}. Hits@10 used for ACC, FWT, +BWT, and REM. Best viewed in color.}
    \label{fig:relation_summary}
\end{figure}

\textbf{Relation sampling evaluations:} Figure~\ref{fig:relation_summary}a summarizes the results of experiments using relation sampling knowledge graph datasets of Table~\ref{tbl:relation_datasets} (WN18RR, FB15K237), where the range of each measure is $[0, 1]$ and larger values are better. Although DGR significantly outperforms other methods in terms of inference (i.e., using ACC and FWT) except for SI in the FB15K237 datasets, there are insights and trade-offs to consider based on other factors.

\begin{small_ind_s_itemize}
    \item Figure~\ref{fig:relation_summary}b shows that DGR has a significantly lower learning speed (based on LCA) than the other methods since a new generative model must be trained in each learning session. PNN had the best learning speed, having significantly better learning speeds than Batch across all datasets and embedding methods. Following close behind were the regularization techniques (L2R and SI), which varied in significant improvements in LCA compared to Batch.
    
    \item Figure~\ref{fig:relation_summary}c indicates that methods with good inference performance also tend to have higher model memory growth (i.e., MS measure); among the methods with significantly better inference performance than Finetune, L2R had the smallest MS followed by SI and DGR.
    
    \item Figure~\ref{fig:relation_summary} indicates that the CKGE methods based on architecture modification (i.e., PNN and CWR) have significantly lower inference performance than Finetune in all experiments. The difference in performance between PNN and the regularization-based methods shows the importance of flexibility over prior concepts for CKGE. Also, CWR's poor inference performance highlights the challenges of directly manipulating the embedding space because, although CWR can learn TE well in isolation, CE is quickly corrupted by the averaging performed to merge embeddings. 
\end{small_ind_s_itemize}

\textbf{Conclusions:} We detailed results obtained with all sampling strategies used for the benchmark knowledge graphs WN18RR and FB15K237. Our results from these extended experiments support the main results presented in the body of the paper. Namely that, while DGR outperforms the other CKGE methods in general, there are trade-offs to consider in terms of learning speed and memory requirements depending on the robotics application.

\end{document}